\documentclass{article}

\PassOptionsToPackage{numbers, sort&compress}{natbib}
\PassOptionsToPackage{hyphens}{url}

\usepackage[preprint]{neurips_2026}

\usepackage[utf8]{inputenc} 
\usepackage[T1]{fontenc}    
\usepackage{hyperref}       
\usepackage{url}            
\usepackage{booktabs}       
\usepackage{amsfonts}       
\usepackage{nicefrac}       
\usepackage{microtype}      
\usepackage{xcolor}         

\usepackage{amsmath, amssymb, amsthm, mathtools}
\usepackage{bm, bbm}
\usepackage{multirow}
\usepackage{makecell}
\usepackage{booktabs}
\usepackage{diagbox}
\usepackage{adjustbox}
\usepackage{float}
\usepackage{wrapfig}
\usepackage{subcaption}
\usepackage[nameinlink,capitalize,noabbrev]{cleveref}
\usepackage[dvipsnames]{xcolor}
\usepackage[most]{tcolorbox}
\usepackage{soul}

\usepackage{newunicodechar}
\usepackage{siunitx}
\usepackage{enumitem}
\usepackage{orcidlink}

\crefname{section}{\S}{\S\S}
\Crefname{section}{\S}{\S\S}
\crefname{paragraph}{\P}{\P\P}
\Crefname{paragraph}{\P}{\P\P}

\title{When2Think: \\Learning When and How Much to Reason}

\author{%
  Jaejun Shim~\orcidlink{0009-0000-1386-5087}$^{1}$ \\
  \texttt{junshim@skku.edu} \\
\And
  HyunJin Kim~\orcidlink{0000-0002-6390-0765}$^{1}$ \\
  \texttt{khyunjin1993@skku.edu} \\
\And
  Young Jin Kim~\orcidlink{0000-0003-2976-3047}$^{2}$~\thanks{Corresponding author} \\
  \texttt{youki@microsoft.com} \\
\And
  JinYeong Bak~\orcidlink{0000-0002-3212-5241}$^{1}$~\footnotemark[1] \\
  \texttt{jy.bak@skku.edu} \\ \\
  $^{1}$Sungkyunkwan University\quad
  $^{2}$Microsoft
}

\begin{document}
\maketitle

\begin{abstract}
Large Reasoning Models (LRMs) often overthink easy problems and underthink hard ones, leading to inefficient computation allocation. 
Existing methods regulate generated computation or select between direct answering and explicit reasoning, but do not jointly control \emph{whether} to reason and \emph{how much} computation to allocate within reasoning. 
We call the resulting difficulty-dependent loss in accuracy under computation reduction the \emph{efficiency tax}. 
We propose \emph{When2Think}, an RLVR-based post-training framework for instance-adaptive computation allocation. 
Its core mechanism, \emph{Instance-level Difficulty-Aware Control (IDAC)}, uses cached reference statistics of success and token cost to modulate a correctness-gated efficiency bonus based on generated token count. 
Importance Sampling supports exploration of \textsc{Think} and \textsc{NoThink}, while Batch-Wise Standardization constructs standardized advantages for critic-free optimization. 
The framework requires neither a learned reward model nor a learned critic, and offline reference caching avoids online reference-model queries during policy updates. 
On AIME24, \emph{When2Think} improves \texttt{Pass@3} by 10.0 percentage points while reducing token usage by 27.9\% relative to the backbone.
We release project at \url{https://github.com/JJunShim/When2Think}.
\end{abstract}

\section{Introduction}
\label{sec:intro}

\begin{figure}[ht]
    \centering
    \begin{subfigure}{0.45\textwidth}
        \centering
        \includegraphics[width=\linewidth]{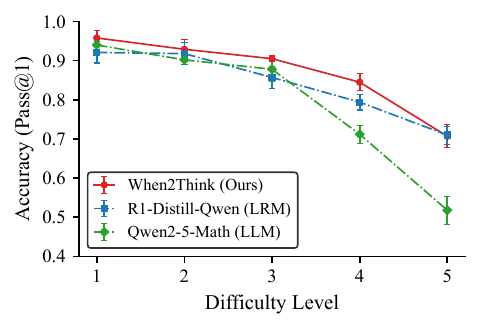}
        \caption{MATH-500 Accuracy}
    \end{subfigure}
    \begin{subfigure}{0.45\textwidth}
        \centering
        \includegraphics[width=\linewidth]{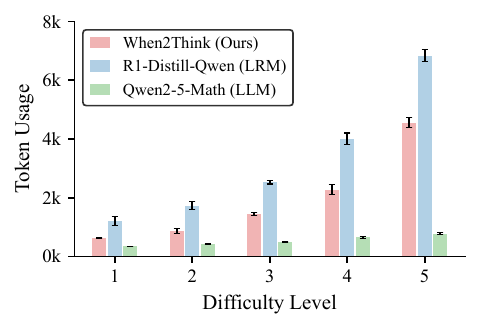}
        \caption{MATH-500 Tokens}
    \end{subfigure}
    \caption{
    \textbf{Performance and efficiency on the MATH-500 benchmark} stratified by problem difficulty (Levels~1--5).
    We compare \emph{When2Think} (\textcolor{BrickRed}{red}), LRM baseline (\textcolor{NavyBlue}{blue}) and LLM baseline (\textcolor{ForestGreen}{green}).
    \textbf{Lines:} Mean \texttt{Pass@1} accuracy $\pm$ standard deviation.
    \textbf{Bars:} Mean tokens per instance $\pm$ standard deviation.
    Statistics computed over five independent sampling runs. 
    \emph{When2Think} mitigates the efficiency tax by reducing token usage on easier instances while preserving hard-instance accuracy.
    }
    \label{fig:problem} 
\end{figure}

The emergence of \emph{large language models (LLMs)}~\citep{NEURIPS2020_1457c0d6, rae2022scalinglanguagemodelsmethods, NEURIPS2022_c1e2faff, JMLR:v24:22-1144, touvron2023llamaopenefficientfoundation} has advanced language understanding and generation across a wide range of tasks.
Beyond fluent text generation, many challenging tasks require reliable multi-step reasoning~\citep{NEURIPS2022_639a9a17, sprague2025to}. 
A prominent direction is to internalize structured reasoning capabilities through \emph{supervised fine-tuning (SFT)} or \emph{reinforcement fine-tuning (RFT)}~\citep{huang-chang-2023-towards, tie2025surveyposttraininglargelanguage, 11640952}, producing \textbf{large reasoning models (LRMs)} that generate coherent reasoning trajectories and achieve strong performance on challenging benchmarks~\citep{ke2025a, XU2025101370, 11267249}.

Despite these advances, LRMs suffer from \emph{reasoning inefficiency}~\citep{feng2025efficient}. 
They often \textbf{overthink} simple tasks by generating redundant reasoning tokens without accuracy gains~\citep{chen2025do, sui2025stop}, yet \textbf{underthink} complex tasks by terminating reasoning prematurely or producing fragmented chains~\citep{wang2025thoughts}. 
Existing efficiency methods reduce computation through computation control or reasoning-mode selection, but can under-allocate computation to harder instances and sacrifice accuracy (\cref{tab:eval_bench}). 
We refer to this difficulty-dependent loss in hard-instance accuracy from reducing computation as the \textbf{efficiency tax}, highlighting the need to control both \emph{whether} and \emph{how much} to reason.

We adopt \emph{dual-process theory} as a conceptual lens for this allocation problem~\citep{Sloman1996-SLOTEC-2, hua-zhang-2022-system}.
\textsc{System~1}-style direct answering provides a low-cost operating point, whereas \textsc{System~2}-style deliberation enables explicit multi-step reasoning~\citep{11267249}.
As shown in \cref{fig:problem}, the LLM baseline uses fewer tokens and performs competitively on easier instances, whereas the LRM baseline benefits more from additional computation as difficulty increases~\citep{shojaee2025the, NEURIPS2025_a797c2d2}.
This contrast motivates \emph{single-model hybrid reasoning}, where a single model adapts computation to the input by selecting between direct answering and explicit deliberation, while varying computation depth within deliberation.

Motivated by these observations, we propose \emph{When2Think}, a post-training framework for difficulty-aware computation control.
At its core, \emph{Instance-level Difficulty-Aware Control (IDAC)} modulates computation-sensitive rewards using reference-estimated difficulty and generated trajectory token count.
Importance-sampled \textsc{Think}/\textsc{NoThink} exploration supports direct answering as an additional low-cost behavior, while IDAC controls how deeply reasons within the \textsc{Think} mode.

We make the following contributions:

\begin{itemize}[leftmargin=*] 
    \item \textbf{Difficulty-stratified analysis.} 
    A difficulty-stratified analysis connects overthinking and underthinking as manifestations of a shared computation-allocation problem, characterizing an \emph{efficiency tax} in which computation savings come at the cost of hard-instance accuracy.
    \item \textbf{Difficulty-aware computation control.} 
    \emph{Instance-level Difficulty-Aware Control (IDAC)} modulates a correctness-gated efficiency bonus using cached reference statistics and the generated token count of each trajectory.
    \item \textbf{Mechanism and behavioral analysis.} 
    The \textsc{Think}-only IDAC+BWS variant outperforms standard RFT on most evaluated benchmarks, demonstrating performance gains within explicit reasoning independently of \textsc{NoThink} selection.
\end{itemize}


\section{Related Work}
\label{sec:related}

We review prior work on reducing the inference cost of large reasoning models (LRMs) through post-training. 
Following recent surveys~\citep{feng2025efficient,sui2025stop}, we organize this literature by the primary control variable: 
(1) \emph{computation control}, which regulates the amount or structure of generated reasoning, and
(2) \emph{hybrid Reasoning mode control}, which determines whether to use direct answering or explicit deliberation. 
Inference-time budgets, stopping rules, and routers provide complementary control over fixed policies.

\paragraph{Computation Control.} 
Global methods apply common efficiency constraints across inputs, including iterative token limits in ThinkPrune~\citep{hou2026thinkprune}, fixed-target step rewards in LASER~\citep{liu2026learn}, and optimization-aware truncation in DLER~\citep{liu2025dlerdoinglengthpenalty}. 
Structural approaches remove redundant reasoning segments through chunk-level distillation (Skip-Thinking~\citep{chen-etal-2025-skip}) or sufficient-prefix extraction (LC-R1~\citep{cheng-etal-2026-optimizing}). 
Adaptive methods derive problem-dependent computation signals from reference statistics or current-policy rollouts, including explicit token budgets in DAST~\citep{shen-etal-2025-dast}, prompt-relative length rewards~\citep{arora2025training}, shortest-correct targets in ShorterBetter~\citep{yi2025shorterbetter}, and dynamically adjusted difficulty-specific thresholds in LASER and difficulty-aware DLER~\citep{liu2026learn,liu2025dlerdoinglengthpenalty}. 
A complementary line studies optimization and credit assignment, showing that computation-sensitive objectives can fail through harmful advantage estimates, entropy collapse, or sparse rollout feedback~\citep{li2026drpo,liu2025dlerdoinglengthpenalty}. 
These studies establish global, structural, and instance-adaptive computation control, but primarily regulate computation within a trajectory.

\paragraph{Reasoning Mode Control.} 
Hybrid methods learn whether an input should receive a direct response or explicit deliberation. 
ARM~\citep{wu2025arm} selects among multiple reasoning formats using diversity-aware reward scaling. Among binary hybrid policies, LHRM~\citep{jiang2025think} separates inter-mode utility comparison from intra-mode response optimization, while ThinkLess~\citep{fang2025thinkless} decouples the control-token and response-token objectives to address length-induced gradient imbalance. 
AdaptThink~\citep{zhang-etal-2025-adaptthink} formulates \textsc{Think}/\textsc{NoThink} selection as a reference-relative constrained objective and uses Importance Sampling to maintain exploration of both modes. 
Inference-time alternatives instead route fixed policies using prompts, confidence estimates, external verifiers, or internal-state probes~\citep{tan2025the}. 
These methods primarily control computation through discrete mode or format decisions, while instance-conditioned control of computation within the selected deliberative mode remains less explored.

\section{Preliminaries}
\label{sec:prel}

Post-training is one effective mechanism for encouraging \textsc{System~2}-style reasoning in LLMs by modifying model parameters, leading to models commonly referred to as \emph{Large Reasoning Models (LRMs)} or \emph{Reasoning Language Models (RLMs)}~\citep{openai2024openaio1card, qwq32b, guo2025deepseek, abdin2025phi4reasoningtechnicalreport, comanici2025gemini25pushingfrontier}. 
We focus on \emph{reinforcement learning (RL)} approaches within the broader paradigm of \emph{learning to reason}~\citep{huang-chang-2023-towards, besta2025reasoninglanguagemodelsblueprint, 11640952}, which directly optimize reasoning policies to enhance both accuracy and efficiency~\citep{trung-etal-2024-reft, ke2025a, XU2025101370, zhang2025surveyreinforcementlearninglarge}. 


\paragraph{Verifiable Rewards.}
\citet{lambert2025tulu} introduce \emph{Reinforcement Learning with Verifiable Rewards (RLVR)}, which replaces learned reward models with deterministic verification, directly aligning the RL objective with task correctness and reducing reward-model bias. 
Prior work~\citep{cobbe2021trainingverifierssolvemath, lightman2024lets} uses verifiers either as \emph{Outcome-supervised Reward Models (ORMs)} for final answers or \emph{Process-supervised Reward Models (PRMs)} for step-level supervision, trading simplicity for annotation cost. 
RLVR leverages outcome-based verification to provide reliable rewards for multi-step reasoning without requiring step-level supervision.




\paragraph{Importance Sampling.} 
To mitigate cold-start bias and avoid premature collapse to \textsc{Think}, we adopt the importance-sampling strategy of AdaptThink~\citep{zhang-etal-2025-adaptthink}. 
During rollout collection, an auxiliary policy $\pi_{\mathrm{IS}}$ selects \textsc{Think} or \textsc{NoThink} with equal probability, after which all tokens are generated by the current policy $\pi_\theta$. 
The collected trajectories are reweighted during optimization to account for the modified mode-selection distribution. 
This mechanism provides balanced hybrid-mode exploration, whereas our contribution focuses on difficulty-aware control of reasoning depth. 
At inference time, $\pi_{\mathrm{IS}}$ is removed and the trained policy selects the mode directly.

\paragraph{Clipped Surrogate Objective.}
We optimize the target policy $\pi_\theta$ using a PPO-style clipped policy gradient objective following AdaptThink~\citep{zhang-etal-2025-adaptthink}, which stabilizes learning of when to invoke explicit reasoning.
Although the objective is defined with respect to trajectories generated by $\pi_\theta$, in practice we optimize using samples collected from an auxiliary policy $\pi_{\mathrm{IS}}$, with importance weighting to ensure consistency with the target policy.

Let $\mathcal{D} = \{(x_i, y_i)\}_{i=1}^{N}$ denote a dataset of $N$ instances. During training, we sample a mini-batch $\mathcal{B} \subset \mathcal{D}$ of size $M$ and reindex its elements as $\{(x_i, y_i)\}_{i=1}^{M}$. For each $x_i \in \mathcal{B}$, we sample $K$ reasoning trajectories, denoted by $\{o_{i,k}\}_{k=1}^{K}$.
The trajectory-level importance ratio:
\begin{equation}
\label{eq:importance_ratio}
\rho_{i,k} = \frac{\pi_\theta(o_{i,k} \mid x_i)}{\pi_{\mathrm{IS}}(o_{i,k} \mid x_i)},
\end{equation}
and the clipped surrogate loss:
\begin{equation}
\label{eq:ppo_at}
\mathcal{L}(\theta) = - \mathbb{E}_{x_i \sim \mathcal{B}, o_{i,k} \sim \pi_{\mathrm{IS}}(\cdot\mid x_i)}
\Big[ \min(\rho_{i,k} \, A_{i,k}, \min(\max(\rho_{i,k}, 1-\epsilon), 1+\epsilon) \, A_{i,k})\Big],
\end{equation}
where $A_{i,k}$ denotes a trajectory-level advantage estimate and $\epsilon$ is the PPO clipping parameter.

\section{When2Think}
\label{sec:method}

\begin{figure*}[tb]
    \centerline{\includegraphics[width=\textwidth]{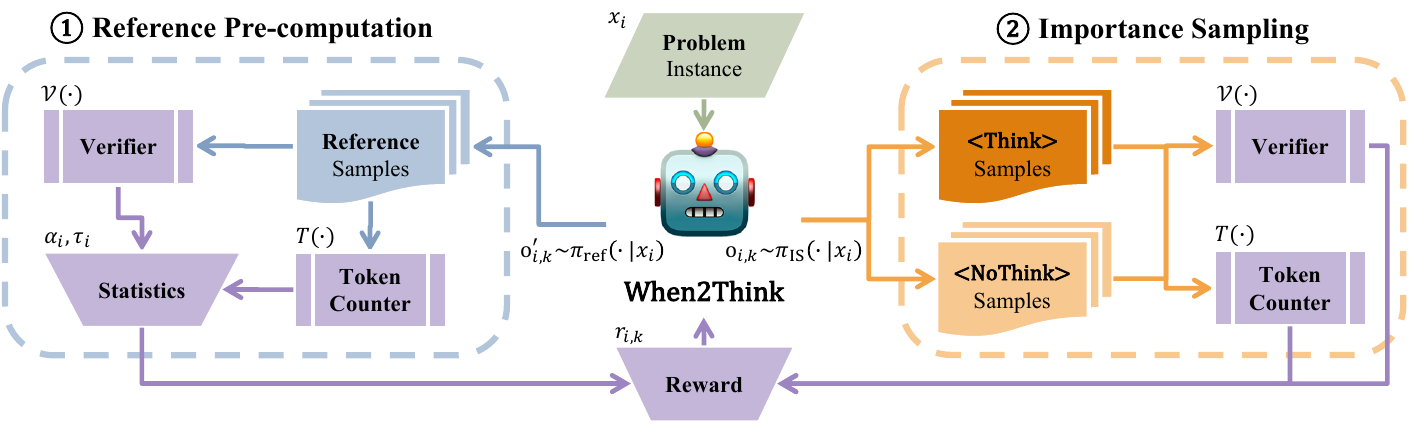}}
    \caption{\textbf{Overview of the \emph{When2Think} RLVR framework} for adaptive hybrid reasoning. 
    The framework comprises two stages. 
    \textbf{(Left) Reference pre-computation:} A reference policy $\pi_{\mathrm{ref}}$ generates trajectories for each input $x_i$ to estimate instance-level difficulty $\alpha_i$ and expected reasoning cost $\tau_i$. 
    \textbf{(Right) Training with Importance Sampling:} An exploration distribution $\pi_{\mathrm{IS}}$ enforces balanced exploration between \textsc{Think} (explicit reasoning) and \textsc{NoThink} (direct answering) by sampling the initial mode token uniformly, while subsequent tokens are generated by the policy $\pi_\theta$. 
    }
    \label{fig:arch} 
\end{figure*}

We propose \emph{When2Think}, a Reinforcement Learning with Verifiable Rewards (RLVR)-based post-training framework for difficulty-aware computation allocation in hybrid reasoning. 
For each input, \emph{When2Think} regulates computation within explicit reasoning while retaining direct answering (\textsc{NoThink}) as an additional low-cost behavior. 
This allocation is guided by cached reference statistics that estimate instance difficulty and expected reasoning cost (\cref{fig:arch}). 
A verifier provides trajectory-level correctness signals, Importance Sampling (IS) supports balanced \textsc{Think}/\textsc{NoThink} exploration, Instance-level Difficulty-Aware Control (IDAC) modulates a correctness-gated efficiency bonus using these statistics and generated token counts, and Batch-Wise Standardization (BWS) constructs standardized advantages for policy optimization. 
Together, these components enable stable optimization without a learned reward model or critic. 
Reference statistics are computed once per epoch and cached, so subsequent policy updates require no online reference-model queries.

\subsection{Reference Statistics}

To estimate instance difficulty and expected reasoning cost without querying the reference model during inner-loop optimization, we pre-compute epoch-wise reference statistics using a reference policy $\pi_{\mathrm{ref}}$ over the dataset $\mathcal{D}$ and cache them for subsequent policy update steps.
For each input $x_i \in \mathcal{D}$, we sample $K$ independent trajectories $o'_{i,k} \sim \pi_{\mathrm{ref}}(\cdot \mid x_i)$.
The final answer is extracted from each trajectory using $g(\cdot)$.
We then define the answer verification function:
\begin{equation}
    \mathcal{V}(x_i,o'_{i,k})
= \mathbb{I}\!\left[\, g(o'_{i,k}) = y_i \,\right],
\end{equation}
where $\mathbb{I}[\cdot]$ equals $1$ if the extracted answer matches the ground-truth $y_i$, and $0$ otherwise. 

\paragraph{Instance Difficulty and Cost.}
To estimate the difficulty and expected reasoning cost of each instance, we compute the reference success rate and average reference token count:

\begin{equation}
\alpha_i
=
\gamma
\left(
\frac{1}{K}
\sum_{k=1}^{K}
\mathcal{V}(x_i,o'_{i,k})
\right)
+
\varepsilon,
\qquad
\tau_i
=
\frac{1}{K}
\sum_{k=1}^{K}
\mathcal{T}(o'_{i,k})
+
\varepsilon.
\end{equation}

Here, $\alpha_i$ is the scaled empirical success rate of the reference policy on instance $x_i$ and serves as a model-relative proxy for instance difficulty, with larger values indicating easier instances. The term $\tau_i$ denotes the average reference token count and provides an instance-specific reference cost scale. The function $\mathcal{T}(\cdot)$ returns the number of generated tokens in a trajectory, while $\gamma \in (0,1)$ controls reward scaling and $\varepsilon > 0$ ensures numerical stability. We use $\gamma=0.9$ in our experiments.

\subsection{Instance-Adaptive Reward}

The adaptive reward combines verifier-based correctness, an instance-level reference baseline, and a correctness-gated efficiency bonus. 
The reference baseline centers trajectory rewards according to the reference policy's empirical success rate for each instance, while IDAC modulates the efficiency bonus using the reasoning mode, generated token count, and cached reference statistics. 
BWS subsequently constructs standardized advantages from the resulting trajectory rewards.

\paragraph{Instance-level Difficulty-Aware Control.}
We introduce \emph{Instance-level Difficulty-Aware Control (IDAC)} to regulate computation according to instance difficulty.
Formally, it is defined:
\begin{equation}
\label{eq:len_decay}
\tilde{\lambda}_{i,k} =
\exp\left( - \frac{\mathcal{T}(o_{i,k}) \, \alpha_i}{\tau_i}\right), \qquad
\lambda_{i,k} = 
\begin{cases}
\tilde{\lambda}_{i,k} & \text{if Think}, \\
1 & \text{if NoThink}.
\end{cases}
\end{equation} 

\begin{wrapfigure}{r}{0.5\textwidth}
    \vspace{-1em}
    \centerline{\includegraphics[width=1\linewidth]{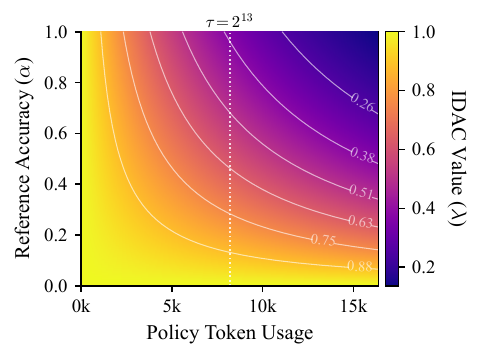}}
    \caption{
    \textbf{Visualization of the Instance-level Difficulty-Aware Control (IDAC) mechanism.} 
    The heatmap shows the IDAC scaling factor applied to \textsc{Think} trajectories as a function of generated token count and reference success rate. 
    For visualization, the reference cost scale is fixed at $\tau=2^{13}$ tokens, half of the maximum \textsc{Think} token limit of $2^{14}$. 
    Higher reference success rates and larger generated token counts produce stronger decay.
    }
    \label{fig:dald} 
    \vspace{-1em}
\end{wrapfigure}


For \textsc{Think} trajectories, $\lambda_{i,k}$ decreases as the generated token count increases relative to the reference cost scale, with stronger decay at higher reference success for a fixed cost scale. 
For \textsc{NoThink} trajectories, $\lambda_{i,k}=1$. 
\cref{fig:dald} illustrates this \textsc{Think}-mode scaling behavior.

\paragraph{Adaptive Reward.}
The final reward combines the answer verification function with an instance-level reference baseline and a correctness-gated efficiency bonus:
\begin{equation}
    r_{i,k}
= \mathcal{V}(x_i,o_{i,k}) \, (1 + \lambda_{i,k}\,\delta) - \alpha_i.
\end{equation}
Here, $\mathcal{V}(x_i,o_{i,k}) \in \{0,1\}$ is the binary correctness signal provided by the verifier, $\lambda_{i,k}$ is the IDAC scaling factor, and $\delta>0$ controls the magnitude of the efficiency bonus. 
The term $\alpha_i$ serves as an instance-specific reference baseline, calibrating trajectory rewards against the reference policy's empirical success rate on $x_i$.
For correct trajectories, \textsc{NoThink} receives the full efficiency bonus. 
For \textsc{Think}, the bonus decreases with generated token count, with stronger decay for higher reference success relative to the reference cost scale. 

\subsection{Batch-Wise Standardized Advantage}

To enable stable trajectory-level credit assignment without a learned critic, we adopt \emph{batch-wise standardization (BWS)}, drawing upon the stabilization insights of \emph{REINFORCE++}~\citep{hu2025reinforcestabilizingcriticfreepolicy}.

Given a mini-batch of $M$ instances, we sample $K$ trajectories for each instance $i$, $\{o_{i,k}\}_{k=1}^{K},$
and compute a scalar adaptive reward $r_{i,k}$ for each trajectory. 
For each trajectory index $k$, we compute the batch-wise mean and standard deviation of rewards across the $M$ instances:
\begin{equation}
\mu_k = \frac{1}{M} \sum_{i=1}^{M} r_{i,k}, 
\qquad
\sigma_k = \sqrt{\frac{1}{M} \sum_{i=1}^{M} (r_{i,k} - \mu_k)^2 }.
\end{equation}

The trajectory-level advantage is defined:
\begin{equation}
A_{i,k} = \frac{r_{i,k} - \mu_k}{\sigma_k + \varepsilon},
\end{equation}
where $\varepsilon > 0$ is a small constant for numerical stability. 
All tokens in a trajectory share the same standardized advantage, yielding a low-variance estimator that naturally supports trajectory-level credit assignment in the PPO-style clipped surrogate optimization described in \cref{sec:prel}.

\section{Experimental Setup}
\label{sec:exp}

We evaluate whether this adaptive control maintains accuracy on difficult instances while reducing unnecessary computation on easier ones. 
Implementation details are provided in \cref{appx:implement}.

\subsection{Training}
We train \emph{When2Think} on a reasoning-capable base model, \emph{R1-distill-1.5B}, a compact language model distilled from \emph{DeepSeek-R1}~\citep{guo2025deepseek} and built on \emph{Qwen2.5-Math-1.5B}~\citep{yang2024qwen25mathtechnicalreportmathematical}.
The model was distilled using approximately 800k curated samples via SFT. 
Training is conducted on the \emph{DeepScaleR} dataset~\citep{tan2026deepscaler}, which contains approximately 40k competition-level mathematics problems with verified solutions.
The dataset aggregates publicly available sources, including AIME (1984--2023), AMC (pre-2023), Omni-MATH~\citep{gao2025omnimath}, and Still~\citep{min2024imitateexploreselfimprovereproduction}, covering algebra, geometry, number theory, and combinatorics.

\subsection{Evaluation}
We evaluate models on mathematical reasoning benchmarks spanning grade-school arithmetic, olympiad challenges, competition-level problems, and undergraduate STEM tasks, including \emph{GSM-Plus}~\citep{li-etal-2024-gsm}, \emph{OlympiadBench}~\citep{he-etal-2024-olympiadbench}, \emph{AIME I/II} (2024--2025), \emph{Minerva}~\citep{NEURIPS2022_18abbeef}, and \emph{MATH-500}~\citep{lightman2024lets}. We additionally evaluate cross-domain transfer on \emph{MMLU-Pro Stratified}~\citep{shi-etal-2025-educationq}, which extends beyond the mathematical post-training distribution.

We evaluate models along two dimensions: \emph{answer accuracy} and \emph{token usage}. Accuracy is measured using sampling-based \texttt{Pass@k}~\citep{chen2021evaluatinglargelanguagemodels}, while efficiency is measured by the average number of generated tokens per instance, computed independently of the \texttt{Pass@k} sampling budget. Final answers are validated using two verifiers, \emph{Math-Verify}~\citep{Kydlicek_Math-Verify_Math_Verification} and \emph{MARIO Eval}~\citep{zhang2024marioevalevaluatemath}; a prediction is considered correct if accepted by either verifier. Further evaluation settings are provided in \cref{appx:eval}.

\subsection{Baselines}

We compare \emph{When2Think} with representative baselines from three categories: (1) base LLMs, (2) standard LRMs trained via SFT or RFT, and (3) efficiency-aware LRMs. 
All evaluated models are built on the Qwen2.5~\citep{qwen2025qwen25technicalreport} family, providing a broadly consistent architectural basis for comparison. 
The base LLMs include the general-purpose \emph{Qwen2.5-Instruct}~\citep{qwen2025qwen25technicalreport} and math-specialized \emph{Qwen2.5-Math-Instruct}~\citep{yang2024qwen25mathtechnicalreportmathematical}. 
For explicit reasoning, we evaluate \emph{R1-Distill-Qwen}~\citep{guo2025deepseek}, which also serves as the backbone of \emph{When2Think}, and \emph{DeepScaleR-Preview}~\citep{tan2026deepscaler}, which is trained on the same \emph{DeepScaleR} dataset. 
Following the control variables identified in \cref{sec:related}, we group efficiency-aware baselines into computation-control and reasoning-mode-control methods. 
Computation-control baselines include \emph{LC-R1}~\citep{cheng-etal-2026-optimizing}, \emph{ThinkPrune}~\citep{hou2026thinkprune}, and \emph{LASER}~\citep{liu2026learn}. 
Reasoning-mode-control baselines include \emph{AdaptThink}~\citep{zhang-etal-2025-adaptthink} and \emph{ThinkLess}~\citep{fang2025thinkless}. 
Among these efficiency-aware baselines, \emph{LASER}, \emph{AdaptThink}, and \emph{ThinkLess} are trained on the same \emph{DeepScaleR} dataset as \emph{When2Think}, enabling more controlled comparisons of their post-training objectives~\citep{tan2026deepscaler,liu2026learn,fang2025thinkless,zhang-etal-2025-adaptthink}.

\section{Experimental Results \& Analysis}
\label{sec:result}

\begin{table*}[t]
    \centering
    \caption{\textbf{Performance comparison on mathematical reasoning benchmarks}.
    We report \texttt{Pass@3} accuracy (\%, higher is better), average tokens per response (\texttt{Tokens}, lower is better) on GSM-Plus, and AIME24. 
    All results use sampling-based \texttt{Pass@k} with matched decoding settings and are reported as mean~$\pm$~standard deviation over five independent sampling runs. 
    \emph{When2Think} is compared with base LLMs, LRM-based baselines, and efficiency-oriented reasoning models. Parentheses indicate differences relative to R1-Distill-Qwen (\textcolor{ForestGreen}{green}: improvement; \textcolor{BrickRed}{red}: degradation).}
    \label{tab:eval_bench}
    \resizebox{1\textwidth}{!}
    {
\begin{tabular}{l|rr|rr|rr}
\toprule
\toprule
\multirowcell{2}{\diagbox{Model}{Benchmark}} & \multicolumn{2}{c|}{\large{GSM-Plus}} & \multicolumn{2}{c|}{\large{AIME24}} & \multicolumn{2}{c}{\large{AIME25}} \\ 
\cline{2-7} 
    & ↑ Pass@3\small{(\%)} & ↓ Tokens\small{(\#)} & ↑ Pass@3\small{(\%)} & ↓ Tokens\small{(\#)} & ↑ Pass@3\small{(\%)} & ↓ Tokens\small{(\#)} \\ 
\midrule
\midrule
Qwen2.5-Math-Instruct   & \num{84.3} \tiny{± \num{0.3}} & \num{360} \tiny{± \num{1.8}} & \num{22.7} \tiny{± \num{1.5}} & \num{1057} \tiny{± \num{57.7}} & \num{19.3} \tiny{± \num{2.8}} & \num{923} \tiny{± \num{64.8}} \\ 
\midrule
\textbf{R1-Distill-Qwen (Base)} & \num{79.4} \tiny{± \num{0.3}} & \num{590} \tiny{± \num{11.6}} & \num{46.0} \tiny{± \num{2.8}} & \num{14195} \tiny{± \num{1026.7}} & \num{32.0} \tiny{± \num{5.1}} & \num{12616} \tiny{± \num{688.4}} \\ 
DeepScaleR-Preview      & \num{85.4} \tiny{± \num{0.3}} & \num{1358} \tiny{± \num{13.4}} & \num{58.0} \tiny{± \num{5.6}} & \num{8473} \tiny{± \num{479.5}} & \num{39.3} \tiny{± \num{6.8}} & \num{8074} \tiny{± \num{288.6}} \\ 
\midrule
LC-R1                   & \makecell[r]{{\num{79.0} \tiny{± \num{0.3}}} \\ {\textcolor{BrickRed}{\small{(- \num{0.4})}}}} & \makecell[r]{\num{546} \tiny{± \num{9.9}} \\ \textcolor{ForestGreen}{\small{(- \num{44})}}}  & \makecell[r]{{\num{36.0} \tiny{± \num{4.9}}} \\ {\textcolor{BrickRed}{\small{(- \num{10.0})}}}} & \makecell[r]{\num{7972} \tiny{± \num{1024.3}} \\ \textcolor{ForestGreen}{\small{(- \num{6223})}}} & \makecell[r]{{\num{32.0} \tiny{± \num{3.8}}} \\ {\textcolor{gray}{\small{(± \num{0})}}}} & \makecell[r]{\num{6971} \tiny{± \num{278.7}} \\ \textcolor{ForestGreen}{\small{(- \num{5645})}}} \\ 
ThinkPrune-iter3k       & \makecell[r]{{\num{81.8} \tiny{± \num{0.3}}} \\ {\textcolor{ForestGreen}{\small{(+ \num{2.4})}}}} & \makecell[r]{\num{609} \tiny{± \num{5.5}} \\ \textcolor{BrickRed}{\small{(+ \num{19})}}} & \makecell[r]{{\num{46.0} \tiny{± \num{4.3}}} \\ {\textcolor{Gray}{\small{(± \num{0})}}}} & \makecell[r]{\num{6236} \tiny{± \num{555.4}} \\ \textcolor{ForestGreen}{\small{(- \num{7959})}}} & \makecell[r]{{\num{36.0} \tiny{± \num{3.7}}} \\ {\textcolor{ForestGreen}{\small{(+ \num{4.0})}}}} & \makecell[r]{\num{5582} \tiny{± \num{273.3}} \\ \textcolor{ForestGreen}{\small{(- \num{7034})}}} \\ 
Laser-DE-L4096          & \makecell[r]{{\num{80.1} \tiny{± \num{0.2}}} \\ {\textcolor{ForestGreen}{\small{(+ \num{0.7})}}}} & \makecell[r]{\num{562} \tiny{± \num{4.1}} \\ \textcolor{ForestGreen}{\small{(- \num{28})}}} & \makecell[r]{\num{45.3} \tiny{± \num{3.8}} \\ \textcolor{BrickRed}{\small{(- \num{0.7})}}} & \makecell[r]{\num{8425} \tiny{± \num{362.3}} \\ \textcolor{ForestGreen}{\small{(- \num{5770})}}} & \makecell[r]{{\num{37.3} \tiny{± \num{8.9}}} \\ {\textcolor{ForestGreen}{\small{(+ \num{5.3})}}}} & \makecell[r]{\num{6971} \tiny{± \num{1024.3}} \\ \textcolor{ForestGreen}{\small{(- \num{5645})}}} \\ 
AdaptThink-delta0.05    & \makecell[r]{\num{83.6} \tiny{± \num{0.2}} \\ \textcolor{ForestGreen}{\small{(+ \num{3.2})}}} & \makecell[r]{\num{716} \tiny{± \num{17.5}} \\ \textcolor{BrickRed}{\small{(+ \num{126})}}} & \makecell[r]{\num{44.7} \tiny{± \num{4.5}} \\ \textcolor{BrickRed}{\small{(- \num{1.3})}}} & \makecell[r]{\num{5806} \tiny{± \num{438.8}} \\ \textcolor{ForestGreen}{\small{(- \num{8389})}}} & \makecell[r]{{\num{30.7} \tiny{± \num{8.0}}} \\ {\textcolor{BrickRed}{\small{(- \num{1.3})}}}} & \makecell[r]{\num{6883} \tiny{± \num{440.2}} \\ \textcolor{ForestGreen}{\small{(- \num{5733})}}} \\ 
Thinkless-RL            & \makecell[r]{\num{85.8} \tiny{± \num{0.2}} \\ \textcolor{ForestGreen}{\small{(+ \num{6.4})}}} & \makecell[r]{\num{1799} \tiny{± \num{16.2}} \\ \textcolor{BrickRed}{\small{(+ \num{1209})}}} & \makecell[r]{\num{46.7} \tiny{± \num{4.1}} \\ \textcolor{ForestGreen}{\small{(+ \num{0.7})}}} & \makecell[r]{\num{11023} \tiny{± \num{209.4}} \\ \textcolor{ForestGreen}{\small{(- \num{3172})}}} & \makecell[r]{{\num{33.3} \tiny{± \num{7.5}}} \\ {\textcolor{ForestGreen}{\small{(+ \num{1.3})}}}} & \makecell[r]{\num{11056} \tiny{± \num{641.7}} \\ \textcolor{ForestGreen}{\small{(- \num{1560})}}} \\ 
\midrule
\textbf{When2Think (Ours)} & \makecell[r]{\num{85.7} \tiny{± \num{0.3}} \\ \textcolor{ForestGreen}{\small{(+ \num{6.3})}}} & \makecell[r]{\num{1052} \tiny{± \num{17.9}} \\ \textcolor{BrickRed}{\small{(+ \num{462})}}} & \makecell[r]{\num{56.0} \tiny{± \num{2.8}} \\ \textcolor{ForestGreen}{\small{(+ \num{10.0})}}} & \makecell[r]{\num{10236} \tiny{± \num{438.0}} \\ \textcolor{ForestGreen}{\small{(- \num{3959})}}} & \makecell[r]{{\num{40.0} \tiny{± \num{0.0}}} \\ {\textcolor{ForestGreen}{\small{(+ \num{8.0})}}}} & \makecell[r]{\num{9549} \tiny{± \num{719.4}} \\ \textcolor{ForestGreen}{\small{(- \num{3067})}}} \\ 
\quad\textbf{w/o IS (IDAC $+$ BWS)} & \makecell[r]{\num{86.8} \tiny{± \num{0.1}} \\ \textcolor{ForestGreen}{\small{(+ \num{7.4})}}} & \makecell[r]{\num{1652} \tiny{± \num{13.2}} \\ \textcolor{BrickRed}{\small{(+ \num{1062})}}} & \makecell[r]{\num{57.3} \tiny{± \num{8.3}} \\ \textcolor{ForestGreen}{\small{(+ \num{11.3})}}} & \makecell[r]{\num{10046} \tiny{± \num{487.9}} \\ \textcolor{ForestGreen}{\small{(- \num{4149})}}} & \makecell[r]{{\num{40.0} \tiny{± \num{3.3}}} \\ {\textcolor{ForestGreen}{\small{(+ \num{8.0})}}}} & \makecell[r]{\num{9846} \tiny{± \num{269.2}} \\ \textcolor{ForestGreen}{\small{(- \num{2770})}}} \\ 
\bottomrule
\bottomrule
\end{tabular}

    }
\end{table*}

We evaluate \emph{When2Think} through three questions covering trade-off, allocation, and mechanism.

\noindent\textbf{RQ1: Accuracy--Efficiency Trade-off under Adaptive Allocation.}
Does \emph{When2Think} improve the accuracy--efficiency trade-off and mitigate the \emph{efficiency tax}?

\noindent\textbf{RQ2: Difficulty-Sensitive Mode and Computation Allocation.}
Does the policy allocate less computation to easy instances and preserve deeper reasoning on hard ones?

\noindent\textbf{RQ3: Mechanism Effects.}
How do \emph{Instance-level Difficulty-Aware Control (IDAC)}, \emph{Batch-Wise Standardization (BWS)}, \emph{Importance Sampling (IS)} enable stable control of reasoning depth?

\subsection{RQ1: Accuracy--Efficiency Trade-off under Adaptive Allocation}
\label{sec:rq1}

We evaluate whether \emph{When2Think} improves the accuracy--efficiency trade-off by reducing unnecessary reasoning on easy instances without inducing under-allocation on hard ones.

\paragraph{The Efficiency Tax.}
Many efficient reasoning methods reduce computation through fixed length constraints, heuristic compression, or discrete mode switching.
While these strategies can suppress redundant reasoning on easy instances, they may incur an \emph{efficiency tax}: token savings are obtained by removing reasoning that is still needed on harder instances.
For example, on AIME24 (\cref{tab:eval_bench}), LC-R1 and AdaptThink reduce inference cost relative to the R1-Distill-Qwen backbone, but their accuracy decreases by \num{10.0} and \num{1.3} points, respectively.
A similar hard-instance under-allocation pattern appears on MATH-500 Level~5 (\cref{tab:eval_math500} in \cref{appx:math}).

\paragraph{Top-Right Accuracy--Efficiency Improvements.}
\cref{fig:benchmarks_comparison} shows the accuracy--efficiency landscape across five benchmarks: GSM-Plus, Minerva, OlympiadBench, AIME24, and AIME25. 
With the token axis reversed, the top-right region represents higher accuracy with fewer tokens. \emph{When2Think} does not necessarily achieve the lowest token usage; routing-only methods such as AdaptThink can reach lower-cost operating points on easier tasks. 
Instead, \emph{When2Think} targets computation reduction while maintaining accuracy on harder instances. 
On AIME24, it reduces token usage by \num{3959} tokens (\num{-27.9}\%) while improving accuracy from \num{46.0}\% to \num{56.0}\% (\num{+10.0} points), showing that adaptive allocation can reduce the efficiency tax rather than merely shift it.

\subsection{RQ2: Difficulty-Sensitive Mode and Computation Allocation}
\label{sec:rq2}

We analyze whether \emph{When2Think} dynamically allocates computation in a \emph{difficulty-aware} manner, balancing \textsc{System~1/2} across benchmarks of varying difficulty.

\paragraph{\textsc{System~1} Preference on Easy Instances.}
On easy MATH-500 (\cref{tab:eval_math500} in \cref{appx:math}) Level~1 problems, the R1-Distill generates an average of \num{1199} tokens despite minimal reasoning demand. 
In contrast, \emph{When2Think} reduces token usage to \num{619} (\num{-580}) while maintaining high accuracy (\num{95.8}\%), indicating a clear preference for \textsc{System~1}-style direct answering when additional deliberation provides limited benefit.
On GSM-Plus, \emph{When2Think} instead allocates additional computation to perturbed inputs, improving Pass@3 by 6.3 percentage points at a higher average token cost.

\paragraph{\textsc{System~2} Engagement on Hard Instances.}
On hard MATH-500 (\cref{tab:eval_math500} in \cref{appx:math}) Level~5 problems, \emph{When2Think} maintains accuracy while reducing token by \num{2276} relative to R1-Distill, indicating more selective yet sustained \textsc{System~2} reasoning.
In contrast, on adversarially perturbed GSM-Plus instances, it deliberately increases computation (\num{462} tokens), yielding a \num{6.3}\% absolute accuracy gain.
On long-horizon tasks such as AIME24/25 (\cref{tab:eval_bench}), \emph{When2Think} maintains sufficient deliberation depth, while compression-based and discrete hybrid methods often truncate reasoning prematurely, leading to poorer performance on hard instances.
When2Think (\cref{appx:mode}) selects \textsc{Think} for all evaluated AIME24 and AIME25 instances. These benchmarks therefore isolate computation control within the reasoning mode rather than active THINK/NOTHINK switching.

\begin{wrapfigure}{r}{0.5\textwidth}
    \vspace{-1.5em}
    \centerline{\includegraphics[width=1\linewidth]{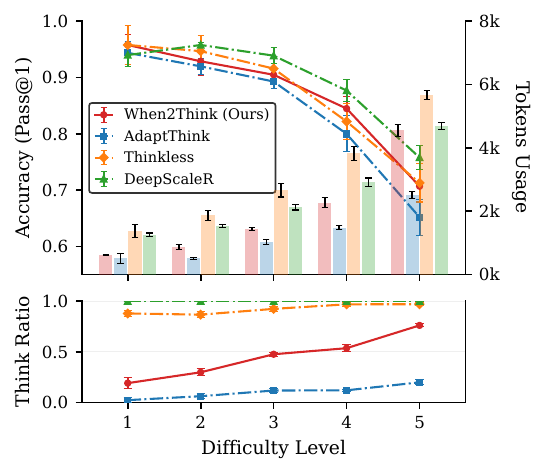}}
    \caption{\textbf{Performance and behavioral analysis on the MATH-500 benchmark} stratified by problem difficulty (Levels~1--5). 
    Results are reported for \emph{When2Think} (\textcolor{BrickRed}{red}), DeepScaleR (\textcolor{ForestGreen}{green}), \emph{Thinkless} (\textcolor{BurntOrange}{orange}), and AdaptThink (\textcolor{NavyBlue}{blue}).
    \textbf{(Top) Accuracy and efficiency}: \texttt{Pass@1} accuracy (lines) and average token usage (bars). 
    \textbf{(Bottom) Think ratio}: Fraction of trajectories that trigger explicit reasoning across difficulty levels. 
    All results are reported as mean $\pm$ standard deviation over five independent sampling runs.
    }
    \label{fig:hybrid} 
    \vspace{-3em}
\end{wrapfigure}

\paragraph{Instance-Level Difficulty-Aware Allocation.}
\emph{When2Think} adapts reasoning behavior at the instance level.
On MATH-500 (\cref{fig:hybrid,appx:mode}), the fraction of \textsc{Think} trajectories increases monotonically with difficulty, from about \num{0.2} at Level~1 to over \num{0.7} at Level~5.
By contrast, baselines exhibit weak or misaligned responses: ThinkLess allocates nearly constant computation, DeepScaleR overthinks easy instances, and AdaptThink under-allocates on hard ones.
This demonstrates that \emph{When2Think} more accurately matches computation to problem difficulty.

\subsection{RQ3: Mechanism Effects}
\label{sec:rq3}

We analyze the mechanisms underlying \emph{When2Think}, focusing on the roles of IDAC, BWS, IS, and model scale. Additional ablations are provided in \cref{appx:ablation}.

\paragraph{IDAC enables depth control.}
Compared with AdaptThink~\citep{zhang-etal-2025-adaptthink}, which primarily learns whether to invoke \textsc{Think}, \emph{When2Think} further controls how deeply to reason once \textsc{Think} is selected. On AIME24 (\cref{tab:eval_bench}), AdaptThink uses fewer tokens (\num{5806}) but drops to \num{44.7}\% Pass@3, below the R1-Distill-Qwen backbone, whereas \emph{When2Think} achieves \num{56.0}\% Pass@3 while still reducing token usage.
Consistent with this, the IS+BWS variant--which retains balanced \textsc{Think}/\textsc{NoThink} exploration but removes IDAC--exhibits substantially worse validation accuracy, particularly on the hard AIME24 benchmark (\cref{fig:is_dald,appx:abl_is_IDAC}).

\paragraph{BWS stabilizes optimization.}
BWS converts IDAC-shaped trajectory rewards into standardized advantages, enabling stable critic-free optimization. Beyond variance reduction, it preserves cross-instance difficulty structure, allowing successful long trajectories on hard instances to receive positive credit. As shown in \cref{app:reward_advantage_landscape,appx:abl_bws}, BWS improves convergence, stabilizes response length and think ratio, and prevents degenerate reasoning behaviors, maintaining a stable think ratio of approximately 0.9 compared to 0.6 without BWS (\cref{fig:training_metrics}).

\paragraph{IS improves efficiency, not accuracy.}
Removing Importance Sampling (while retaining IDAC and BWS, \cref{tab:eval_bench}) yields a strong variant that outperforms DeepScaleR-Preview on most benchmarks, indicating that the primary accuracy gains stem from IDAC and BWS. However, IS shifts the policy toward a more efficient operating point: compared with the w/o IS variant, the full \emph{When2Think} model substantially reduces GSM-Plus token usage (\num{1652} $\rightarrow$ \num{1052} tokens) while maintaining competitive accuracy. This suggests that IS primarily promotes effective \textsc{NoThink} usage on easier instances, rather than serving as the main source of accuracy gains (\cref{appx:abl_is_IDAC}).


\section{Discussion}
\label{sec:disc}

\subsection{From Compression to Computation Allocation}

Our results suggest that reasoning efficiency is better understood as an \emph{instance-adaptive computation-allocation problem} than as uniform trajectory compression. Reducing computation can improve efficiency on easier instances but incur an \emph{efficiency tax} when harder instances are under-allocated and lose accuracy (\cref{tab:eval_math,tab:eval_math500}). Effective reasoning therefore requires adapting computation to individual inputs rather than uniformly shortening responses. Difficulty-stratified evaluation is important for revealing where computation is reduced, preserved, or expanded.

\subsection{Mode Selection and Within-\textsc{Think} Control}

Hybrid reasoning involves two distinct but coupled decisions: whether to invoke explicit reasoning and how much computation to allocate after it is invoked. Our results show that these behaviors can vary independently. On MATH-500 (\cref{fig:hybrid,tab:eval_math500}), the \textsc{Think} ratio increases with difficulty, while on AIME24 and AIME25 (\cref{appx:mode}) the policy selects \textsc{Think} for every evaluated instance. Thus, computation control within \textsc{Think} remains relevant even when mode switching is inactive.

The ablations further clarify the roles of the training components. The \textsc{Think}-only IDAC+BWS variant remains strong without \textsc{NoThink} exploration (\cref{tab:eval_math}), indicating that the primary performance gains do not depend on hybrid routing alone. IDAC provides instance-conditioned incentives for generated computation within \textsc{Think}, while BWS converts the resulting reward structure into standardized advantages for critic-free optimization. Importance sampling plays a complementary role by maintaining coverage of both modes and enabling lower-cost \textsc{NoThink} behavior where it is effective (\cref{tab:eval_math,appx:ablation}). The full hybrid and \textsc{Think}-only policies therefore represent different accuracy--computation operating points rather than a strict performance ordering.

\subsection{Scope and Limitations}

Allocation behavior varies with model scale and the reference policy (\cref{appx:prob,appx:math}). At 7B, \emph{When2Think} reduces computation on MATH-500 but allocates more computation on AIME24 and AIME25, showing that effective allocation does not always mean shorter responses.

IDAC uses reference success and token usage as proxies for difficulty and computation cost, so its allocation behavior may depend on the reference policy. Generated token count is also only an operational proxy for inference cost. Finally, our evaluation focuses on verifiable tasks, leaving generalization to open-ended reasoning as future work.

\section{Conclusion}
\label{sec:conc}

We formulate efficient reasoning as an \emph{instance-adaptive computation-allocation problem} involving two coupled decisions: whether to invoke explicit reasoning and how much computation to allocate once it is invoked. We introduce \emph{When2Think}, an RLVR-based post-training framework that combines importance-sampled mode exploration, IDAC-based computation control, and BWS-based cross-instance credit assignment. IDAC uses cached reference success and token-cost statistics to shape a correctness-gated efficiency bonus within \textsc{Think}, while BWS enables stable, critic-free optimization without a learned reward model. 

Our results show difficulty-sensitive mode allocation on mixed-difficulty inputs and effective within-\textsc{Think} computation control when explicit deliberation remains necessary. The \textsc{Think}-only IDAC+BWS variant demonstrates that computation control within explicit reasoning can improve performance independently of \textsc{NoThink} exploration, while the full hybrid policy additionally provides a lower-cost operating point on suitable inputs. Together, these findings establish mode selection and within-mode computation as distinct but coupled components of efficient reasoning, shifting the objective from uniform trajectory shortening toward instance-adaptive computation allocation.

\bibliographystyle{unsrtnat}
\bibliography{references}

\clearpage
\appendix

\section{Implementation details}
\label{appx:implement}

\subsection{Reproducibility}
All experiments prioritize transparency and reproducibility. 
Each model is trained once with a fixed seed of $42$ and evaluated over five runs using distinct sampling seeds. 
Hyperparameter configurations (\cref{tab:params}) follow prior studies on reasoning alignment for fair comparison.

\begin{table}[H]
    \centering
    \caption{\textbf{Hyperparameter} for training configuration.}
    \label{tab:params}
    \resizebox{0.5\textwidth}{!}
    {
        \begin{tabular}{l|rr}
\toprule
\toprule
Hyperparameter & 1.5B & 7B \\
\midrule
\midrule
\multicolumn{3}{l}{\textit{Optimization}} \\
\midrule
Total Episodes    & \num{40315} & \num{40315} \\
Total Epochs      & \num{1} & \num{1} \\
Total Steps       & \num{314} & \num{628} \\
Batch Size        & \num{128} & \num{64} \\
Learning Rate     & \num{3e-6} & \num{7e-7} \\
Dynamic Batch     & True & False \\
\midrule
\multicolumn{3}{l}{\textit{Generation \& Architecture}} \\
\midrule
Max Prompt Length                     & \num{1024} & \num{1024} \\
Max \textsc{Think} Response Length    & \num{16384} & \num{16384} \\
Max \textsc{NoThink} Response Length  & \num{4096} & \num{4096} \\
Reference Sampling ($K$)              & \num{16} & \num{16} \\
Importance Sampling ($K$)             & \num{16} & \num{16} \\
Temperature                           & \num{.6} & \num{.6} \\
Top-p                                 & \num{.95} & \num{.95} \\
Random Seed                           & \num{42} & \num{42} \\
Weight Type                           & float32 & float32 \\
\midrule
\multicolumn{3}{l}{\textit{Reward Configuration (When2Think)}} \\
\midrule
Reference Scaler ($\gamma$)  & \num{0.9} & \num{0.9} \\
\textsc{NoThink} Ratio                & \num{.5} & \num{.5} \\
Bonus ($\delta$)  & \num{.05} & \num{.05} \\
\bottomrule
\bottomrule
\end{tabular}

    }
\end{table}

\subsection{Infrastructure}
Training the 1.5B model required approximately 70 GPU-hours on 2$\times$ NVIDIA H100 80GB GPUs, while the 7B model required approximately 120 GPU-hours.
Evaluation across all reported benchmarks required approximately 20 GPU-hours on 2$\times$ NVIDIA A100 80GB GPUs.
Training throughput was maximized using the \emph{verl} framework~\citep{10.1145/3689031.3696075} with Distributed Data Parallel (DDP) and FlashAttention-2 integration.
Inference and evaluation employed the \emph{vLLM} engine~\citep{10.1145/3600006.3613165}, which reduces memory overhead and latency for long chain-of-thought generation.
Comprehensive software and hardware specifications are summarized in \cref{tab:env}.

\begin{table}[H]
    \centering
    \caption{\textbf{Computational environment and software stack} used in experiments.}
    \label{tab:env}
    \resizebox{0.65\textwidth}{!}
    {
        \begin{tabular}{l|rr}
\toprule
\toprule
Environment & Train & Test \\
\midrule
\midrule
\multicolumn{3}{l}{\textit{Hardware Accelerator}} \\
\midrule

Accelerator         & NVIDIA H100 (80GB HBM3) & NVIDIA A100 (80GB) \\
\midrule
\multicolumn{3}{l}{\textit{System Environment}} \\
\midrule
OS Kernel    & Linux-5.15.0-161-generic & Linux-5.4.0-173-generic \\
Python              & CPython-3.11.14 & CPython 3.11.14 \\
CUDA                & 13.0 & 12.8 \\
\midrule
\multicolumn{3}{l}{\textit{Core Frameworks}} \\
\midrule
\multirow{3}{*}{\raggedright Frameworks} & PyTorch-2.9.0 & PyTorch-2.9.0 \\
                    & Transformers-4.57.1 & Transformers-4.57.3 \\
                    & verl-0.2.0.dev & vLLM-0.11.2 \\  
\bottomrule
\bottomrule
\end{tabular}

    }
\end{table}

\section{Evaluation Setup}
\label{appx:eval}

\subsection{Inference and Verification Protocol}
To ensure a rigorous and reproducible evaluation, we standardize inference-time generation parameters and the answer verification pipeline across all models.

\paragraph{Generation Configuration.}
We adopt decoding strategies tailored to the characteristics of each model class while keeping them fixed within each category for fair comparison.
For general-purpose LLMs, we use a \texttt{temperature} of $0.7$ and \texttt{top-p} of $0.8$ to encourage diverse solution exploration.
For Large Reasoning Models (LRMs), we apply a more deterministic configuration with a \texttt{temperature} of $0.6$ and \texttt{top-p} of $0.95$, which stabilizes long-horizon reasoning while preserving sufficient diversity in intermediate steps.
This distinction follows common practice in prior work and accounts for the differing sensitivity of LLMs and LRMs to sampling noise during reasoning, while all generation settings are held constant across benchmarks to ensure fair comparison.

To ensure reliable answer extraction and automated evaluation, all models are instructed to follow a strict output format using the system prompt shown in \cref{box:sys_prompt}, which is instantiated using each model`s native chat template.

\begin{tcolorbox}[enhanced, breakable, title=System Prompt for Mathematical Reasoning, label={box:sys_prompt}]
Math assistant. Provide a step-by-step solution in English using \LaTeX. Final answer in \verb|\boxed{...}|.
\end{tcolorbox}

\paragraph{Answer Extraction and Verification.}
We report \texttt{Pass@k} accuracy based on answers extracted from the final \verb|\boxed{...}| tag.
To ensure robustness to formatting variations and symbolic ambiguity, we employ a \textbf{Dual Math Verifier} pipeline that integrates \emph{Math-Verify}~\citep{Kydlicek_Math-Verify_Math_Verification} and \emph{MARIO Eval}~\citep{zhang2024marioevalevaluatemath}.
Math-Verify performs primary extraction and symbolic equivalence checking, while MARIO Eval provides complementary semantic verification using type-specific Computer Algebra Systems (CAS).
The resulting unified binary correctness signal is used consistently as both the evaluation metric and the terminal reward for policy optimization.

\subsection{Dual Math Verifier}
\label{app:dual_math_verifier}

Accurate and robust verification is essential for evaluating mathematical reasoning and for constructing reliable reward signals during policy optimization.
Because mathematically correct answers can be expressed in multiple symbolic or numerical forms, exact string matching is often insufficient.
We therefore employ a dual-verifier pipeline that evaluates semantic equivalence between model-generated answers and ground-truth solutions.

\paragraph{Math-Verify.}
We use Math-Verify~\citep{Kydlicek_Math-Verify_Math_Verification} as a format-agnostic verifier for mathematical reasoning tasks.
Given a model-generated response, Math-Verify extracts the candidate answer using prioritized parsing rules that support diverse output formats, including LaTeX expressions and plain text.
The extracted answer is then normalized and converted into a symbolic representation, enabling comparison with the ground-truth solution beyond exact string matching.
Verification is performed through symbolic equivalence checking and numerical evaluation with tolerance, allowing equivalent mathematical expressions to be recognized even when their surface forms differ.

\paragraph{MARIO Eval.}
We additionally adopt MARIO Eval~\citep{zhang2024marioevalevaluatemath} to provide type-aware verification across diverse mathematical answer formats.
MARIO Eval first identifies the answer type, such as a scalar, algebraic expression, matrix, equation, or inequality, using rule-based parsing and optional language-model assistance when type inference is ambiguous.
It then applies type-specific equivalence checks, primarily relying on symbolic computation through a computer algebra system and numerical evaluation with tolerance.
When symbolic verification is inconclusive due to structural ambiguity or natural-language formulations, MARIO Eval can optionally invoke a math-capable language model as a fallback judge.

\subsection{Benchmarks}
We evaluate models on a diverse suite of mathematical reasoning benchmarks spanning foundational arithmetic, intermediate competition problems, olympiad-level challenges, and undergraduate STEM tasks.
Together, these benchmarks cover a broad difficulty spectrum, enabling fine-grained analysis of both reasoning accuracy and computational efficiency across problem complexities.


\paragraph{GSM-Plus.}
GSM-Plus~\citep{li-etal-2024-gsm} is a medium-scale benchmark of approximately 10.5k problems that extends GSM-8K~\citep{cobbe2021trainingverifierssolvemath} with systematically perturbed variants.
It introduces controlled modifications, including paraphrasing, numerical changes, distractor insertion, and reasoning-structure perturbations, enabling evaluation of robustness under increased reasoning complexity and distribution shift.
Although GSM-Plus contains a small number of instances annotated with \texttt{None} answers for unanswerable or incomplete cases, we exclude them to maintain a consistent answer-extraction and verification protocol.
This filtering yields a final evaluation set of approximately 9k problems.

\paragraph{MATH-500.}
The MATH-500 benchmark~\citep{lightman2024lets} is a held-out evaluation subset derived from the original MATH~\citep{NEURIPS2021_be83ab3e} dataset, preserving its structure while avoiding contamination from the PRM-800K~\citep{lightman2024lets} training pipeline. 
The full MATH dataset spans seven subjects--Prealgebra, Algebra, Number Theory, Counting and Probability, Geometry, Intermediate Algebra, and Precalculus--and assigns each problem a difficulty level from 1 to 5 based on AoPS conventions, ranging from introductory AMC 8--style problems to AIME-level items. 
To prevent overfitting due to inclusion of test-split MATH problems in PRM-800K, 500 uniformly sampled problems were isolated to form MATH-500. 
This subset retains the subject distribution and difficulty profile of the original test split, providing a leakage-free benchmark for evaluating advanced mathematical reasoning.

\paragraph{Minerva.}
The Undergraduate-Level STEM Problems, also known as the Minerva benchmark and referred to as OCW Courses in \citet{NEURIPS2022_18abbeef}, consists of 272 self-contained problems collected from MIT OpenCourseWare materials across multiple STEM subjects, including Solid-State Chemistry, Information and Entropy, Differential Equations, and Special Relativity. 
Each problem provides a clearly delineated final answer and is automatically verifiable, either numerically (191 problems) or symbolically via SymPy (81 problems), emphasizing multi-step reasoning while excluding formal proofs or open-ended short answers. 
This benchmark is specifically designed to evaluate scientific and quantitative reasoning capabilities in large language models at the undergraduate level, bridging the gap between standard K-12 mathematics benchmarks and more advanced domain-specific problem-solving tasks.


\paragraph{OlympiadBench-Math.}
OlympiadBench~\citep{he-etal-2024-olympiadbench} is a high-difficulty, bilingual benchmark comprising 8,476 problems across multiple STEM domains in both multimodal and text-only formats. 
For evaluation of mathematical reasoning, only text-only math problems are used to isolate multi-step reasoning performance. 
Each problem includes an expert-verified solution, enabling reliable assessment of model reasoning. 
Experiments are conducted on a curated subset of 674 open-ended, text-only English problems from the Math Competition track of OlympiadBench.

\paragraph{AIME.} 
The American Invitational Mathematics Examination (AIME) is a selective competition featuring integer-answer problems ranging from 000 to 999. These problems are substantially more challenging than standard AMC questions, requiring multi-step reasoning, careful case analysis, and deeper understanding of algebra, combinatorics, number theory, and geometry. 
For evaluation, both administrations (AIME I and AIME II) are combined to form a set of 30 problems. 
The 2024 and 2025 edition is used to avoid overlap with the DeepScaleR dataset, which includes material only through 2023.



\clearpage

\subsection{THINK/NOTHINK Trajectory Formats} 
\label{appx:hybrid_implementation} 
We implement hybrid reasoning using two trajectory formats: \textsc{Think} and \textsc{NoThink}. Let $\langle\mathrm{BOT}\rangle$ and $\langle\mathrm{EOT}\rangle$ denote the backbone-specific begin-of-thinking and end-of-thinking delimiters, respectively. The concrete delimiters and chat-template representation may vary across model families. 

A \textsc{Think} trajectory contains an explicit reasoning block: 
\[ \textsc{Think}: [x_i] + \langle\mathrm{BOT}\rangle + [\text{reasoning}] + \langle\mathrm{EOT}\rangle + [\text{answer}]. \] 

A \textsc{NoThink} trajectory bypasses the dedicated reasoning block by placing the end-of-thinking delimiter immediately after the begin-of-thinking delimiter: 
\[ \textsc{NoThink}: [x_i] + \langle\mathrm{BOT}\rangle + \langle\mathrm{EOT}\rangle + [\text{answer}]. \] 

Thus, \textsc{NoThink} is conceptually implemented as an empty reasoning block. In our experiments, the concrete formatting follows the native reasoning template of the DeepSeek-R1~\citep{guo2025deepseek}. For a different backbone, the corresponding model-specific reasoning delimiters or chat-template controls should be used. 

During rollout collection, the importance-sampling exploration policy constructs \textsc{Think} and \textsc{NoThink} trajectories with equal probability. After the mode-specific prefix is constructed, all remaining tokens are generated autoregressively by the current policy. 
At inference time, the exploration policy is removed, and the trained policy directly determines whether to continue generating within the reasoning block or terminate the block and enter the answer channel. 
Operationally, \textsc{NoThink} denotes bypassing the dedicated reasoning block.

\subsection{Thoughts \& Steps}
\citet{lu2025retrosearchexploringuntakenpaths} propose a hierarchical decomposition of reasoning trajectories by distinguishing between \emph{thoughts} and \emph{steps}.

Given a question $q \in \mathcal{Q}$, a reasoning model $\mathcal{M}$ produces a reasoning trajectory $T$ and a final answer $a$, denoted as $(T, a) := \mathcal{M}(q)$. 
The trajectory $T$ is organized as a sequence of higher-level thoughts, each of which consists of multiple intermediate reasoning steps.

\paragraph{Thoughts ($s^\tau$).} 
    A reasoning trajectory $T$ is represented as an ordered sequence of thoughts:
    \begin{equation}
        T := \langle s^1, s^2, \dots, s^{\tau}, \dots, s^m \rangle ,
    \end{equation}
    where each $s^\tau$ corresponds to a contiguous segment of reasoning that follows a coherent line of argument or problem-solving strategy. 
    Transitions between consecutive thoughts, $s^{\tau} \rightarrow s^{\tau+1}$, are identified using discourse-level linguistic markers (e.g., ``Alternatively'', ``Wait''; see \cref{box:thoughts}), which heuristically indicate a shift in reasoning direction.

\begin{tcolorbox}[enhanced, breakable, title={Discourse Markers Used for Segmenting Thoughts}, label={box:thoughts}]
But, Wait, Alternatively, However, Hmm, Hmmm, Not sure, Going back, Backtrack, Trace back, Another
\end{tcolorbox}

\paragraph{Steps ($s^\tau_k$).}
    Each thought $s^\tau$ is further decomposed into a sequence of intermediate steps:
    \begin{equation}
        s^\tau := \langle s^\tau_1, s^\tau_2, \dots, s^\tau_{k_\tau} \rangle .
    \end{equation}
    The superscript $\tau$ indexes the thought, while the subscript $k$ indexes the step within that thought. 
    In practice, steps correspond to fine-grained inference units and are delimited using formatting cues in the generated trace, such as double newline characters (\texttt{\textbackslash n\textbackslash n}).

Under this formulation, a reasoning trajectory admits the following hierarchical structure:
\begin{equation}
    T = \langle 
    \langle s^1_1, \dots, s^1_{k_1} \rangle,
    \dots,
    \langle s^\tau_1, \dots, s^\tau_{k_\tau} \rangle,
    \dots
    \rangle .
\end{equation}

This hierarchical decomposition is used solely as an analysis tool to describe the structure of reasoning trajectories.

\clearpage
\subsection{Accuracy-Cost Efficiency}
We introduce \emph{Accuracy-Cost Efficiency (\texttt{ACE})}, a metric that measures how efficiently a model`s generation leads to a correct final answer under hybrid \textsc{System~1/2} reasoning.
\texttt{ACE} extends prior efficiency metrics, including \emph{Outcome Efficiency}~\citep{chen2025do}, \emph{VT}~\citep{cheng-etal-2026-optimizing}, and \emph{Efficiency}~\citep{li2025thinkbenchevaluatingthinkingefficiency}, to hybrid reasoning settings. 
Unlike these measures, \texttt{ACE} accommodates optional or partially expressed reasoning and does not assume that correct answers are embedded within a single continuous reasoning trace.

Let $N$ denote the number of test instances. 
For each instance $i$, let $\sigma_i \in \{0,1\}$ indicate whether the final answer is correct, $\tau_i$ denote the total number of generated tokens, and $\hat{\tau}_i$ denote the position of the first token where the correct answer appears.
Let $m_i^{\textrm{Think}} \in \{0,1\}$ indicate whether the model produces explicit reasoning.

In hybrid reasoning scenarios, the correct answer may not appear explicitly in the reasoning trace or may be produced without any explicit reasoning. 
In such cases, we set $\hat{\tau}_i = \tau_i$, corresponding to maximal efficiency under \texttt{ACE}, since no redundant reasoning tokens are generated beyond the final answer.

\paragraph{Per-Instance Efficiency.}
Based on the above definitions, the per-instance efficiency is computed:
\begin{equation}
\label{eq:delta}
\eta_i
= 1 
- m_i^{\textrm{Think}} 
\, \mathbb{I}[\hat{\tau}_i < \tau_i] 
\, \left(1 - \frac{\hat{\tau}_i}{\tau_i}\right),
\end{equation}
where:
\begin{itemize}
    \item $\eta_i = 1$ if either no explicit reasoning is generated ($m_i^{\textrm{Think}} = 0$) or the correct answer first appears at the end of the trajectory ($\hat{\tau}_i = \tau_i$),
    \item $\eta_i < 1$ if explicit reasoning continues beyond the first correct token, decreasing proportionally to the fraction of redundant reasoning tokens.
\end{itemize}

\paragraph{Overall \texttt{ACE} Score.}
Finally, the overall \texttt{ACE} score is computed:
\begin{equation}
\label{eq:hoe_clean}
\xi_{\mathrm{ACE}}
= \frac{1}{N} \sum_{i=1}^{N} \, \sigma_i \, \eta_i .
\end{equation}

This score is bounded, $\xi_{\mathrm{ACE}} \in [0,1]$, and is used solely as an evaluation metric.  
Incorrect predictions contribute zero to the score, while correct predictions achieve maximal efficiency if no redundant reasoning occurs.  
If explicit reasoning continues beyond the earliest correct answer, the score decreases proportionally to the fraction of redundant reasoning tokens, $1 - \hat{\tau}_i / \tau_i$.

\clearpage
\section{Additional Experimental Results}
\label{appx:add_results}

\subsection{Analysis of Token Distribution}
\label{appx:token}
\begin{figure}[H]
    \centering
    \includegraphics[width=\textwidth]{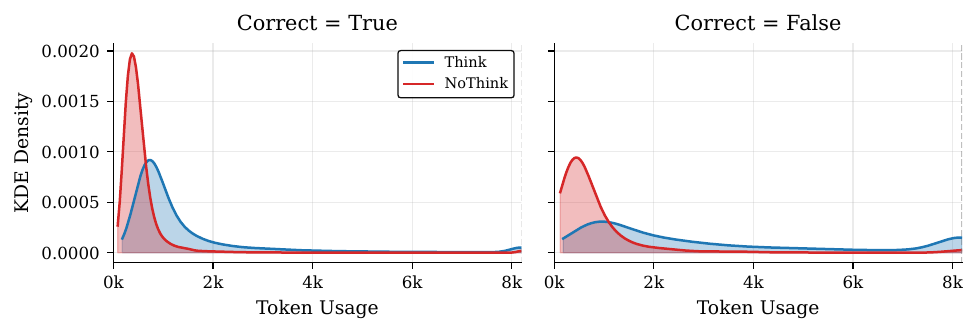}
    \caption{\textbf{When2Think token usage distributions by reasoning mode.} 
    Kernel Density Estimation (KDE) of generated token counts for \textsc{Think} (\textcolor{NavyBlue}{blue}) and \textsc{NoThink} (\textcolor{BrickRed}{red}), stratified by response correctness. 
    \textbf{Left:} Correct responses--\textsc{NoThink} peaks at low token usage. 
    \textbf{Right:} Incorrect responses--\textsc{Think} exhibits a broader distribution with a long tail, indicating higher token consumption during failed reasoning.}
    \label{fig:token_dist}
\end{figure}

\cref{fig:token_dist} shows the distributions of token usage by reasoning mode and response correctness. 
For correct instances (Left), \textsc{NoThink} peaks sharply at low token counts, reflecting \textsc{System~1} efficiency, whereas \textsc{Think} consumes more tokens even when correct, revealing overthinking. 
For incorrect instances (Right), \textsc{Think} exhibits a heavy-tailed distribution reaching the context limit, demonstrating the high cost of failed \textsc{System~2} reasoning. 

\subsection{Reasoning Mode Performance}
\label{appx:mode}

\begin{table}[H]
    \centering
    \caption{\textbf{Detailed Performance by Reasoning Mode on GSM-Plus.} 
    The table reports \texttt{Pass@3} accuracy (\%), average token usage ($\downarrow$~\texttt{Tokens}), and step counts ($\downarrow$~\texttt{Steps}) for the \textsc{Think} mode, alongside accuracy, token usage, and selection ratio for the \textsc{NoThink} mode. 
    All metrics are reported as mean~$\pm$~standard deviation over five runs.}
    \label{tab:eval_hybrid_gsmplus}
    \resizebox{0.8\textwidth}{!}
    {
        \begin{tabular}{l|rrr|rrr}
\toprule
\toprule
\multirowcell{3}{\diagbox{Model}{Benchmark}} & \multicolumn{6}{c}{\large{GSM-Plus}} \\ 
\cline{2-7}     & \multicolumn{3}{c|}{\textsc{Think}} & \multicolumn{3}{c}{\textsc{NoThink}} \\
\cline{2-7}     & ↑ Pass@3\small{(\%)} & ↓ Tokens\small{(\#)} & ↓ Steps\small{(\#)} & ↑ Pass@3\small{(\%)} & ↓ Tokens\small{(\#)} & Ratio\small{(\%)} \\ 
\midrule
\midrule
\multicolumn{7}{l}{\large{\textit{\textbf{1.5B Models}}}} \\ 
\midrule
AdaptThink-delta0.05 & \num{67.4} \tiny{± \num{1.6}} & \num{1807} \tiny{± \num{59}} & \num{10} \tiny{± \num{1}} & \num{85.9} \tiny{± \num{0.2}} & \num{561} \tiny{± \num{16}} & \num{87.5} \tiny{± \num{0.1}} \\
Thinkless-RL        & \num{84.5} \tiny{± \num{0.2}} & \num{1872} \tiny{± \num{30}} & \num{14} \tiny{± \num{0}} & \num{93.7} \tiny{± \num{0.8}} & \num{1378} \tiny{± \num{62}} & \num{14.7} \tiny{± \num{0.5}} \\
\textbf{When2Think (Ours)}     & \num{80.8} \tiny{± \num{0.3}} & \num{1401} \tiny{± \num{27}} & \num{21} \tiny{± \num{1}} & \num{93.0} \tiny{± \num{0.5}} & \num{538} \tiny{± \num{15}} & \num{40.3} \tiny{± \num{0.3}} \\
\bottomrule
\bottomrule
\end{tabular}
    }
\end{table}

\begin{table}[H]
    \centering
    \caption{\textbf{Detailed Performance by Reasoning Mode on Minerva.} 
    The table reports \texttt{Pass@3} accuracy (\%), average token usage ($\downarrow$~\texttt{Tokens}), and step counts ($\downarrow$~\texttt{Steps}) for the \textsc{Think} mode, alongside accuracy, token usage, and selection ratio for the \textsc{NoThink} mode. 
    All metrics are reported as mean~$\pm$~standard deviation over five runs.}
    \label{tab:eval_hybrid_minerva}
    \resizebox{0.8\textwidth}{!}
    {
        \begin{tabular}{l|rrr|rrr}
\toprule
\toprule
\multirowcell{3}{\diagbox{Model}{Benchmark}} & \multicolumn{6}{c}{\large{Minerva}} \\ 
\cline{2-7}     & \multicolumn{3}{c|}{\textsc{Think}} & \multicolumn{3}{c}{\textsc{NoThink}} \\
\cline{2-7}     & ↑ Pass@3\small{(\%)} & ↓ Tokens\small{(\#)} & ↓ Steps\small{(\#)} & ↑ Pass@3\small{(\%)} & ↓ Tokens\small{(\#)} & Ratio\small{(\%)} \\ 
\midrule
\midrule
\multicolumn{7}{l}{\large{\textit{\textbf{1.5B Models}}}} \\ 
\midrule
AdaptThink-delta0.05 & \num{36.6} \tiny{± \num{1.6}} & \num{3012} \tiny{± \num{285}} & \num{34} \tiny{± \num{6}} & \num{57.3} \tiny{± \num{1.6}} & \num{1154} \tiny{± \num{96}} & \num{58.7} \tiny{± \num{2.6}} \\
Thinkless-RL        & \num{52.5} \tiny{± \num{2.0}} & \num{4982} \tiny{± \num{200}} & \num{74} \tiny{± \num{7}} & \num{47.4} \tiny{± \num{10.1}} & \num{6487} \tiny{± \num{1603}} & \num{5.5} \tiny{± \num{1.1}} \\
\textbf{When2Think (Ours)}     & \num{50.3} \tiny{± \num{2.5}} & \num{3282} \tiny{± \num{82}} & \num{92} \tiny{± \num{3}} & \num{78.8} \tiny{± \num{4.3}} & \num{811} \tiny{± \num{192}} & \num{9.0} \tiny{± \num{0.6}} \\
\bottomrule
\bottomrule
\end{tabular}
    }
\end{table}

\begin{table}[H]
    \centering
    \caption{\textbf{Detailed Performance by Reasoning Mode on OlympiadBench.} 
    The table reports \texttt{Pass@3} accuracy (\%), average token usage ($\downarrow$~\texttt{Tokens}), and step counts ($\downarrow$~\texttt{Steps}) for the \textsc{Think} mode, alongside accuracy, token usage, and selection ratio for the \textsc{NoThink} mode. 
    All metrics are reported as mean~$\pm$~standard deviation over five runs.}
    \label{tab:eval_hybrid_olympiad}
    \resizebox{0.8\textwidth}{!}
    {
        \begin{tabular}{l|rrr|rrr}
\toprule
\toprule
\multirowcell{3}{\diagbox{Model}{Benchmark}} & \multicolumn{6}{c}{\large{OlympiadBench-Math}} \\ 
\cline{2-7}     & \multicolumn{3}{c|}{\textsc{Think}} & \multicolumn{3}{c}{\textsc{NoThink}} \\
\cline{2-7}     & ↑ Pass@3\small{(\%)} & ↓ Tokens\small{(\#)} & ↓ Steps\small{(\#)} & ↑ Pass@3\small{(\%)} & ↓ Tokens\small{(\#)} & Ratio\small{(\%)} \\ 
\midrule
\midrule
\multicolumn{7}{l}{\large{\textit{\textbf{1.5B Models}}}} \\ 
\midrule
AdaptThink-delta0.05 & \num{41.5} \tiny{± \num{2.1}} & \num{7223} \tiny{± \num{215}} & \num{69} \tiny{± \num{4}} & \num{62.5} \tiny{± \num{1.3}} & \num{2003} \tiny{± \num{54}} & \num{80.5} \tiny{± \num{0.9}} \\
Thinkless-RL        & \num{63.6} \tiny{± \num{0.9}} & \num{6839} \tiny{± \num{116}} & \num{117} \tiny{± \num{4}} & \num{59.4} \tiny{± \num{9.0}} & \num{9785} \tiny{± \num{979}} & \num{4.0} \tiny{± \num{0.7}} \\
\textbf{When2Think (Ours)}     & \num{57.5} \tiny{± \num{1.2}} & \num{6635} \tiny{± \num{57}} & \num{204} \tiny{± \num{3}} & \num{90.5} \tiny{± \num{2.1}} & \num{1086} \tiny{± \num{93}} & \num{14.7} \tiny{± \num{0.5}} \\
\bottomrule
\bottomrule
\end{tabular}
    }
\end{table}

\begin{table}[H]
    \centering
    \caption{\textbf{Detailed Performance by Reasoning Mode on AIME24.} 
    The table reports \texttt{Pass@3} accuracy (\%), average token usage ($\downarrow$~\texttt{Tokens}), and step counts ($\downarrow$~\texttt{Steps}) for the \textsc{Think} mode, alongside accuracy, token usage, and selection ratio for the \textsc{NoThink} mode. 
    All metrics are reported as mean~$\pm$~standard deviation over five runs.}
    \label{tab:eval_hybrid_aime24}
    \resizebox{0.8\textwidth}{!}
    {
        \begin{tabular}{l|rrr|rrr}
\toprule
\toprule
\multirowcell{3}{\diagbox{Model}{Benchmark}} & \multicolumn{6}{c}{\large{AIME24}} \\ 
\cline{2-7}     & \multicolumn{3}{c|}{\textsc{Think}} & \multicolumn{3}{c}{\textsc{NoThink}} \\
\cline{2-7}     & ↑ Pass@3\small{(\%)} & ↓ Tokens\small{(\#)} & ↓ Steps\small{(\#)} & ↑ Pass@3\small{(\%)} & ↓ Tokens\small{(\#)} & Ratio\small{(\%)} \\ 
\midrule
\midrule
\multicolumn{7}{l}{\large{\textit{\textbf{1.5B Models}}}} \\ 
\midrule
AdaptThink-delta0.05 & \num{44.0} \tiny{± \num{11.9}} & \num{8121} \tiny{± \num{678}} & \num{338} \tiny{± \num{72}} & \num{47.6} \tiny{± \num{12.8}} & \num{3227} \tiny{± \num{552}} & \num{48.0} \tiny{± \num{11.7}} \\
Thinkless-RL        & \num{47.7} \tiny{± \num{4.6}} & \num{10861} \tiny{± \num{245}} & \num{221} \tiny{± \num{31}} & \num{0.0} \tiny{± \num{0.0}} & \num{18897} \tiny{± \num{1606}} & \num{2.0} \tiny{± \num{1.8}} \\
\textbf{When2Think (Ours)}     & \num{56.0} \tiny{± \num{2.8}} & \num{10236} \tiny{± \num{438}} & \num{399} \tiny{± \num{27}} & -- & -- & \num{0.0} \tiny{± \num{0.0}} \\
\bottomrule
\bottomrule
\end{tabular}
    }
\end{table}

\begin{table}[H]
    \centering
    \caption{\textbf{Detailed Performance by Reasoning Mode on AIME25.} 
    The table reports \texttt{Pass@3} accuracy (\%), average token usage ($\downarrow$~\texttt{Tokens}), and step counts ($\downarrow$~\texttt{Steps}) for the \textsc{Think} mode, alongside accuracy, token usage, and selection ratio for the \textsc{NoThink} mode. 
    All metrics are reported as mean~$\pm$~standard deviation over five runs.}
    \label{tab:eval_hybrid_aime25}
    \resizebox{0.8\textwidth}{!}
    {
        \begin{tabular}{l|rrr|rrr}
\toprule
\toprule
\multirowcell{3}{\diagbox{Model}{Benchmark}} & \multicolumn{6}{c}{\large{AIME25}} \\ 
\cline{2-7}     & \multicolumn{3}{c|}{\textsc{Think}} & \multicolumn{3}{c}{\textsc{NoThink}} \\
\cline{2-7}     & ↑ Pass@3\small{(\%)} & ↓ Tokens\small{(\#)} & ↓ Steps\small{(\#)} & ↑ Pass@3\small{(\%)} & ↓ Tokens\small{(\#)} & Ratio\small{(\%)} \\ 
\midrule
\midrule
\multicolumn{7}{l}{\large{\textit{\textbf{1.5B Models}}}} \\ 
\midrule
AdaptThink-delta0.05 & \num{29.7} \tiny{± \num{12.6}} & \num{9114} \tiny{± \num{828}} & \num{345} \tiny{± \num{30}} & \num{32.7} \tiny{± \num{3.3}} & \num{3000} \tiny{± \num{585}} & \num{36.7} \tiny{± \num{4.1}} \\
Thinkless-RL        & \num{35.2} \tiny{± \num{8.9}} & \num{10667} \tiny{± \num{729}} & \num{210} \tiny{± \num{35}} & \num{0.0} \tiny{± \num{0.0}} & \num{19497} \tiny{± \num{2328}} & \num{4.7} \tiny{± \num{5.1}} \\
\textbf{When2Think (Ours)}     & \num{40.0} \tiny{± \num{0.0}} & \num{9549} \tiny{± \num{719}} & \num{424} \tiny{± \num{65}} & -- & -- & \num{0.0} \tiny{± \num{0.0}} \\
\bottomrule
\bottomrule
\end{tabular}
    }
\end{table}

\subsection{Performance Across Benchmarks}
\label{appx:math}

\begin{figure}[H]
    \centering
    \includegraphics[width=\textwidth]{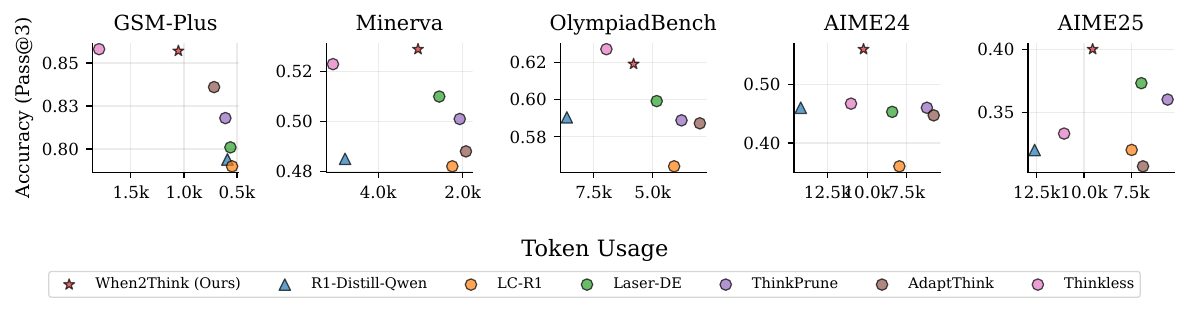}
    \caption{\textbf{Accuracy-Efficiency trade-off across reasoning benchmarks.}
    Performance of \textbf{\emph{When2Think} (Ours, $\star$)} is compared with the base model \textbf{R1-Distill-Qwen (Triangle)} on five benchmarks of varying difficulty: GSM-Plus, Minerva, OlympiadBench, AIME24, and AIME25.
    The x-axis shows \textbf{Token Usage} (reversed; further right indicates fewer tokens/more efficient), and the y-axis shows \textbf{Pass@3 Accuracy}.
    Across benchmarks, \emph{When2Think} points generally occupy the top-right region relative to the baseline, indicating higher accuracy with lower token usage.}
    \label{fig:benchmarks_comparison}
\end{figure}

\begin{table}[H]
    \centering
    \caption{\textbf{Extended performance comparison on reasoning benchmarks.} 
    The table reports \texttt{Pass@3} accuracy (\%), average tokens per response (↓ \texttt{Tokens}), and Accuracy-Cost Efficiency (↑ \texttt{ACE}) on GSM-Plus, Minerva, OlympiadBench-Math, AIME24, and AIME25 under a unified evaluation protocol. 
    Results are obtained using a sampling-based \texttt{Pass@k} procedure with matched decoding settings and are reported as mean~$\pm$~standard deviation over five runs. 
    Values in parentheses indicate differences relative to R1-Distill-Qwen (\textcolor{ForestGreen}{green}: improvement; \textcolor{BrickRed}{red}: degradation).}
    \label{tab:eval_math}
    \begin{adjustbox}{angle=270, max height=\textheight}
        \resizebox{0.9\textheight}{!}
        {
            \begin{tabular}{l|rrr|rrr|rrr|rrr|rrr}
\toprule
\toprule
\multirowcell{2}{\diagbox{Model}{Benchmark}} & \multicolumn{3}{c|}{\large{GSM-Plus}} & \multicolumn{3}{c|}{\large{Minerva}} & \multicolumn{3}{c|}{\large{OlympiadBench-Math}} & \multicolumn{3}{c|}{\large{AIME24}} & \multicolumn{3}{c}{\large{AIME25}} \\ 
\cline{2-16} 
    & \multicolumn{1}{c}{↑ Pass@3\small{(\%)}}
    & \multicolumn{1}{c}{↓ Tokens\small{(\#)}}
    & \multicolumn{1}{c|}{↑ ACE\small{(\%)}}
    & \multicolumn{1}{c}{↑ Pass@3\small{(\%)}}
    & \multicolumn{1}{c}{↓ Tokens\small{(\#)}}
    & \multicolumn{1}{c|}{↑ ACE\small{(\%)}}
    & \multicolumn{1}{c}{↑ Pass@3\small{(\%)}}
    & \multicolumn{1}{c}{↓ Tokens\small{(\#)}}
    & \multicolumn{1}{c|}{↑ ACE\small{(\%)}}
    & \multicolumn{1}{c}{↑ Pass@3\small{(\%)}}
    & \multicolumn{1}{c}{↓ Tokens\small{(\#)}}
    & \multicolumn{1}{c|}{↑ ACE\small{(\%)}}
    & \multicolumn{1}{c}{↑ Pass@3\small{(\%)}}
    & \multicolumn{1}{c}{↓ Tokens\small{(\#)}}
    & \multicolumn{1}{c}{↑ ACE\small{(\%)}}
 \\ 
\midrule
\midrule
\multicolumn{16}{l}{\large{\textit{\textbf{1.5B Models}}}} \\ 
\midrule
Qwen2.5-Instruct        & \num{73.6} \tiny{± \num{0.3}} & \num{346} \tiny{± \num{2.9}} & \multicolumn{1}{c|}{--} & \num{30.1} \tiny{± \num{1.5}} & \num{716} \tiny{± \num{53.2}} & \multicolumn{1}{c|}{--} & \num{29.6} \tiny{± \num{0.9}} & \num{976} \tiny{± \num{31.0}} & \multicolumn{1}{c|}{--} & \num{8.0} \tiny{± \num{3.8}} & \num{1373} \tiny{± \num{343.1}} & \multicolumn{1}{c|}{--} & \num{1.3} \tiny{± \num{3.0}} & \num{995} \tiny{± \num{163.4}} & \multicolumn{1}{c}{--} \\ 
Qwen2.5-Math-Instruct   & \num{84.3} \tiny{± \num{0.3}} & \num{360} \tiny{± \num{1.8}} & \multicolumn{1}{c|}{--} & \num{44.8} \tiny{± \num{0.5}} & \num{671} \tiny{± \num{14.7}} & \multicolumn{1}{c|}{--} & \num{53.2} \tiny{± \num{0.8}} & \num{842} \tiny{± \num{8.3}} & \multicolumn{1}{c|}{--} & \num{22.7} \tiny{± \num{1.5}} & \num{1057} \tiny{± \num{57.7}} & \multicolumn{1}{c|}{--} & \num{19.3} \tiny{± \num{2.8}} & \num{923} \tiny{± \num{64.8}} & \multicolumn{1}{c}{--} \\ 
\midrule
\textbf{R1-Distill-Qwen (Base)} & \num{79.4} \tiny{± \num{0.3}} & \num{590} \tiny{± \num{11.6}} & \num{79.2} \tiny{± \num{0.3}} & \num{48.5} \tiny{± \num{1.9}} & \num{4782} \tiny{± \num{124.1}} & \num{36.4} \tiny{± \num{2.1}} & \num{58.8} \tiny{± \num{2.0}} & \num{8634} \tiny{± \num{107.4}} & \num{39.6} \tiny{± \num{1.6}} & \num{46.0} \tiny{± \num{2.8}} & \num{14195} \tiny{± \num{1026.7}} & \num{26.2} \tiny{± \num{3.7}} & \num{32.0} \tiny{± \num{5.1}} & \num{12616} \tiny{± \num{688.4}} & \num{17.7} \tiny{± \num{3.1}} \\ 
DeepScaleR-Preview      & \num{85.4} \tiny{± \num{0.3}} & \num{1358} \tiny{± \num{13.4}} & \num{83.7} \tiny{± \num{0.1}} & \num{55.4} \tiny{± \num{1.1}} & \num{4130} \tiny{± \num{56.4}} & \num{43.7} \tiny{± \num{1.0}} & \num{62.8} \tiny{± \num{0.8}} & \num{5215} \tiny{± \num{43.3}} & \num{42.9} \tiny{± \num{0.5}} & \num{58.0} \tiny{± \num{5.6}} & \num{8473} \tiny{± \num{479.5}} & \num{38.4} \tiny{± \num{4.4}} & \num{39.3} \tiny{± \num{6.8}} & \num{8074} \tiny{± \num{288.6}} & \num{27.0} \tiny{± \num{5.7}} \\ 
\midrule
LC-R1                   & \makecell[r]{{\num{79.0} \tiny{± \num{0.3}}} \\ {\textcolor{BrickRed}{\small{(- \num{0.4})}}}} & \makecell[r]{\num{546} \tiny{± \num{9.9}} \\ \textcolor{ForestGreen}{\small{(- \num{44})}}} & \makecell[r]{{\num{79.0} \tiny{± \num{0.3}}} \\ {\textcolor{BrickRed}{\small{(- \num{0.2})}}}} & \makecell[r]{{\num{48.2} \tiny{± \num{1.7}}} \\ {\textcolor{BrickRed}{\small{(- \num{0.3})}}}} & \makecell[r]{\num{2248} \tiny{± \num{71.8}} \\ \textcolor{ForestGreen}{\small{(- \num{2534})}}} & \makecell[r]{{\num{42.7} \tiny{± \num{1.8}}} \\ {\textcolor{ForestGreen}{\small{(+ \num{6.3})}}}} & \makecell[r]{{\num{55.5} \tiny{± \num{1.1}}} \\ {\textcolor{BrickRed}{\small{(- \num{3.3})}}}} & \makecell[r]{\num{4092} \tiny{± \num{113.0}} \\ \textcolor{ForestGreen}{\small{(- \num{4542})}}} & \makecell[r]{{\num{39.9} \tiny{± \num{1.1}}} \\ {\textcolor{ForestGreen}{\small{(+ \num{0.3})}}}} & \makecell[r]{{\num{36.0} \tiny{± \num{4.9}}} \\ {\textcolor{BrickRed}{\small{(- \num{10.0})}}}} & \makecell[r]{\num{7972} \tiny{± \num{1024.3}} \\ \textcolor{ForestGreen}{\small{(- \num{6223})}}} & \makecell[r]{{\num{23.0} \tiny{± \num{3.4}}} \\ {\textcolor{BrickRed}{\small{(- \num{3.2})}}}} & \makecell[r]{{\num{32.0} \tiny{± \num{3.8}}} \\ {\textcolor{gray}{\small{(± \num{0})}}}} & \makecell[r]{\num{6971} \tiny{± \num{278.7}} \\ \textcolor{ForestGreen}{\small{(- \num{5645})}}} & \makecell[r]{{\num{18.5} \tiny{± \num{2.7}}} \\ {\textcolor{ForestGreen}{\small{(+ \num{0.8})}}}} \\ 
ThinkPrune-iter3k       & \makecell[r]{{\num{81.8} \tiny{± \num{0.3}}} \\ {\textcolor{ForestGreen}{\small{(+ \num{2.4})}}}} & \makecell[r]{\num{609} \tiny{± \num{5.5}} \\ \textcolor{BrickRed}{\small{(+ \num{19})}}} & \makecell[r]{{\num{79.3} \tiny{± \num{0.3}}} \\ {\textcolor{ForestGreen}{\small{(+ \num{0.1})}}}} & \makecell[r]{{\num{50.1} \tiny{± \num{1.9}}} \\ {\textcolor{ForestGreen}{\small{(+ \num{1.6})}}}} & \makecell[r]{\num{2072} \tiny{± \num{62.2}} \\ \textcolor{ForestGreen}{\small{(- \num{2710})}}} & \makecell[r]{{\num{32.9} \tiny{± \num{1.5}}} \\ {\textcolor{BrickRed}{\small{(- \num{3.5})}}}} & \makecell[r]{{\num{58.6} \tiny{± \num{0.7}}} \\ {\textcolor{ForestGreen}{\small{(+ \num{0.2})}}}} & \makecell[r]{\num{3781} \tiny{± \num{66.5}} \\ \textcolor{ForestGreen}{\small{(- \num{4853})}}} & \makecell[r]{{\num{35.7} \tiny{± \num{0.6}}} \\ {\textcolor{BrickRed}{\small{(- \num{3.9})}}}} & \makecell[r]{{\num{46.0} \tiny{± \num{4.3}}} \\ {\textcolor{gray}{\small{(± \num{0})}}}} & \makecell[r]{\num{6236} \tiny{± \num{555.4}} \\ \textcolor{ForestGreen}{\small{(- \num{7959})}}} & \makecell[r]{{\num{27.0} \tiny{± \num{2.0}}} \\ {\textcolor{ForestGreen}{\small{(+ \num{0.8})}}}} & \makecell[r]{{\num{36.0} \tiny{± \num{3.7}}} \\ {\textcolor{ForestGreen}{\small{(+ \num{4.0})}}}} & \makecell[r]{\num{5582} \tiny{± \num{273.3}} \\ \textcolor{ForestGreen}{\small{(- \num{7034})}}} & \makecell[r]{{\num{21.5} \tiny{± \num{1.9}}} \\ {\textcolor{ForestGreen}{\small{(+ \num{3.8})}}}} \\ 
Laser-DE-L4096          & \makecell[r]{{\num{80.1} \tiny{± \num{0.2}}} \\ {\textcolor{ForestGreen}{\small{(+ \num{0.7})}}}} & \makecell[r]{\num{562} \tiny{± \num{4.1}} \\ \textcolor{ForestGreen}{\small{(- \num{28})}}} & \makecell[r]{\num{79.4} \tiny{± \num{0.3}} \\ \textcolor{ForestGreen}{\small{(+ \num{0.2})}}} & \makecell[r]{\num{51.0} \tiny{± \num{1.2}} \\ \textcolor{ForestGreen}{\small{(+ \num{2.5})}}} & \makecell[r]{\num{2556} \tiny{± \num{105.5}} \\ \textcolor{ForestGreen}{\small{(- \num{2226})}}} & \makecell[r]{\num{38.7} \tiny{± \num{0.7}} \\ \textcolor{ForestGreen}{\small{(+ \num{2.3})}}} & \makecell[r]{\num{59.9} \tiny{± \num{1.1}} \\ \textcolor{ForestGreen}{\small{(+ \num{1.1})}}} & \makecell[r]{\num{4830} \tiny{± \num{43.4}} \\ \textcolor{ForestGreen}{\small{(- \num{3804})}}} & \makecell[r]{\num{39.1} \tiny{± \num{1.0}} \\ \textcolor{BrickRed}{\small{(- \num{0.5})}}} & \makecell[r]{\num{45.3} \tiny{± \num{3.8}} \\ \textcolor{BrickRed}{\small{(- \num{0.7})}}} & \makecell[r]{\num{8425} \tiny{± \num{362.3}} \\ \textcolor{ForestGreen}{\small{(- \num{5770})}}} & \makecell[r]{\num{27.0} \tiny{± \num{2.6}} \\ \textcolor{ForestGreen}{\small{(+ \num{0.8})}}} & \makecell[r]{{\num{37.3} \tiny{± \num{8.9}}} \\ {\textcolor{ForestGreen}{\small{(+ \num{5.3})}}}} & \makecell[r]{\num{6971} \tiny{± \num{1024.3}} \\ \textcolor{ForestGreen}{\small{(- \num{5645})}}} & \makecell[r]{{\num{24.1} \tiny{± \num{6.6}}} \\ {\textcolor{ForestGreen}{\small{(+ \num{6.4})}}}} \\ 
AdaptThink-delta0.05    & \makecell[r]{\num{83.6} \tiny{± \num{0.2}} \\ \textcolor{ForestGreen}{\small{(+ \num{3.2})}}} & \makecell[r]{\num{716} \tiny{± \num{17.5}} \\ \textcolor{BrickRed}{\small{(+ \num{126})}}} & \makecell[r]{\num{83.5} \tiny{± \num{0.2}} \\ \textcolor{ForestGreen}{\small{(+ \num{4.3})}}} & \makecell[r]{\num{48.8} \tiny{± \num{1.3}} \\ \textcolor{ForestGreen}{\small{(+ \num{0.3})}}} & \makecell[r]{\num{1925} \tiny{± \num{110.9}} \\ \textcolor{ForestGreen}{\small{(- \num{2857})}}} & \makecell[r]{\num{48.0} \tiny{± \num{1.5}} \\ \textcolor{ForestGreen}{\small{(+ \num{11.6})}}} & \makecell[r]{\num{58.4} \tiny{± \num{0.8}} \\ \textcolor{BrickRed}{\small{(- \num{0.4})}}} & \makecell[r]{\num{3004} \tiny{± \num{76.2}} \\ \textcolor{ForestGreen}{\small{(- \num{5630})}}} & \makecell[r]{\num{56.9} \tiny{± \num{0.9}} \\ \textcolor{ForestGreen}{\small{(+ \num{17.3})}}} & \makecell[r]{\num{44.7} \tiny{± \num{4.5}} \\ \textcolor{BrickRed}{\small{(- \num{1.3})}}} & \makecell[r]{\num{5806} \tiny{± \num{438.8}} \\ \textcolor{ForestGreen}{\small{(- \num{8389})}}} & \makecell[r]{\num{40.1} \tiny{± \num{4.9}} \\ \textcolor{ForestGreen}{\small{(+ \num{13.9})}}} & \makecell[r]{{\num{30.7} \tiny{± \num{8.0}}} \\ {\textcolor{BrickRed}{\small{(- \num{1.3})}}}} & \makecell[r]{\num{6883} \tiny{± \num{440.2}} \\ \textcolor{ForestGreen}{\small{(- \num{5733})}}} & \makecell[r]{{\num{25.1} \tiny{± \num{5.7}}} \\ {\textcolor{ForestGreen}{\small{(+ \num{7.4})}}}} \\ 
Thinkless-RL            & \makecell[r]{\num{85.8} \tiny{± \num{0.2}} \\ \textcolor{ForestGreen}{\small{(+ \num{6.4})}}} & \makecell[r]{\num{1799} \tiny{± \num{16.2}} \\ \textcolor{BrickRed}{\small{(+ \num{1209})}}} & \makecell[r]{\num{82.7} \tiny{± \num{0.2}} \\ \textcolor{ForestGreen}{\small{(+ \num{3.5})}}} & \makecell[r]{\num{52.3} \tiny{± \num{1.4}} \\ \textcolor{ForestGreen}{\small{(+ \num{3.8})}}} & \makecell[r]{\num{5067} \tiny{± \num{166.0}} \\ \textcolor{BrickRed}{\small{(+ \num{285})}}} & \makecell[r]{\num{44.3} \tiny{± \num{1.4}} \\ \textcolor{ForestGreen}{\small{(+ \num{7.9})}}} & \makecell[r]{\num{63.4} \tiny{± \num{1.1}} \\ \textcolor{ForestGreen}{\small{(+ \num{4.6})}}} & \makecell[r]{\num{6960} \tiny{± \num{76.2}} \\ \textcolor{ForestGreen}{\small{(- \num{1674})}}} & \makecell[r]{\num{47.5} \tiny{± \num{1.5}} \\ \textcolor{ForestGreen}{\small{(+ \num{7.9})}}} & \makecell[r]{\num{46.7} \tiny{± \num{4.1}} \\ \textcolor{ForestGreen}{\small{(+ \num{0.7})}}} & \makecell[r]{\num{11023} \tiny{± \num{209.4}} \\ \textcolor{ForestGreen}{\small{(- \num{3172})}}} & \makecell[r]{\num{30.4} \tiny{± \num{4.4}} \\ \textcolor{ForestGreen}{\small{(+ \num{4.2})}}} & \makecell[r]{{\num{33.3} \tiny{± \num{7.5}}} \\ {\textcolor{ForestGreen}{\small{(+ \num{1.3})}}}} & \makecell[r]{\num{11056} \tiny{± \num{641.7}} \\ \textcolor{ForestGreen}{\small{(- \num{1560})}}} & \makecell[r]{{\num{21.4} \tiny{± \num{4.7}}} \\ {\textcolor{ForestGreen}{\small{(+ \num{3.7})}}}} \\ 
\midrule
\textbf{When2Think (Ours)} & \makecell[r]{\num{85.7} \tiny{± \num{0.3}} \\ \textcolor{ForestGreen}{\small{(+ \num{6.3})}}} & \makecell[r]{\num{1052} \tiny{± \num{17.9}} \\ \textcolor{BrickRed}{\small{(+ \num{462})}}} & \makecell[r]{\num{80.7} \tiny{± \num{0.4}} \\ \textcolor{ForestGreen}{\small{(+ \num{1.5})}}} & \makecell[r]{\num{52.9} \tiny{± \num{2.1}} \\ \textcolor{ForestGreen}{\small{(+ \num{4.4})}}} & \makecell[r]{\num{3058} \tiny{± \num{58.6}} \\ \textcolor{ForestGreen}{\small{(- \num{1724})}}} & \makecell[r]{\num{40.6} \tiny{± \num{1.7}} \\ \textcolor{ForestGreen}{\small{(+ \num{4.2})}}} & \makecell[r]{\num{62.4} \tiny{± \num{0.9}} \\ \textcolor{ForestGreen}{\small{(+ \num{3.6})}}} & \makecell[r]{\num{5909} \tiny{± \num{56.4}} \\ \textcolor{ForestGreen}{\small{(- \num{2725})}}} & \makecell[r]{\num{43.9} \tiny{± \num{1.2}} \\ \textcolor{ForestGreen}{\small{(+ \num{4.3})}}} & \makecell[r]{\num{56.0} \tiny{± \num{2.8}} \\ \textcolor{ForestGreen}{\small{(+ \num{10.0})}}} & \makecell[r]{\num{10236} \tiny{± \num{438.0}} \\ \textcolor{ForestGreen}{\small{(- \num{3959})}}} & \makecell[r]{\num{32.1} \tiny{± \num{1.8}} \\ \textcolor{ForestGreen}{\small{(+ \num{5.9})}}} & \makecell[r]{{\num{40.0} \tiny{± \num{0.0}}} \\ {\textcolor{ForestGreen}{\small{(+ \num{8.0})}}}} & \makecell[r]{\num{9549} \tiny{± \num{719.4}} \\ \textcolor{ForestGreen}{\small{(- \num{3067})}}} & \makecell[r]{{\num{24.6} \tiny{± \num{1.1}}} \\ {\textcolor{ForestGreen}{\small{(+ \num{6.9})}}}} \\ 
\midrule
\multicolumn{16}{l}{\large{\textit{\textbf{7B Models}}}} \\ 
\midrule
Qwen2.5-Instruct        & \num{88.8} \tiny{± \num{0.2}} & \num{342} \tiny{± \num{0.2}} & \multicolumn{1}{c|}{--} & \num{54.0} \tiny{± \num{1.4}} & \num{634} \tiny{± \num{1.4}} & \multicolumn{1}{c|}{--} & \num{52.3} \tiny{± \num{1.4}} & \num{819} \tiny{± \num{11.0}} & \multicolumn{1}{c|}{--} & \num{18.0} \tiny{± \num{3.8}} & \num{1087} \tiny{± \num{107.0}} & \multicolumn{1}{c|}{--} & \num{20.7} \tiny{± \num{7.2}} & \num{937} \tiny{± \num{13.5}} & \multicolumn{1}{c}{--} \\ 
Qwen2.5-Math-Instruct   & \num{89.1} \tiny{± \num{0.2}} & \num{421} \tiny{± \num{3.1}} & \multicolumn{1}{c|}{--} & \num{51.6} \tiny{± \num{1.2}} & \num{833} \tiny{± \num{11.3}} & \multicolumn{1}{c|}{--} & \num{55.8} \tiny{± \num{0.5}} & \num{1110} \tiny{± \num{13.4}} & \multicolumn{1}{c|}{--} & \num{24.0} \tiny{± \num{3.8}} & \num{1614} \tiny{± \num{140.6}} & \multicolumn{1}{c|}{--} & \num{22.0} \tiny{± \num{4.5}} & \num{1512} \tiny{± \num{71.5}} & \multicolumn{1}{c}{--} \\ 
\midrule
R1-Distill-Qwen & \num{87.6} \tiny{± \num{0.1}} & \num{521} \tiny{± \num{1.1}} & \num{87.5} \tiny{± \num{0.1}} & \num{63.6} \tiny{± \num{1.1}} & \num{3517} \tiny{± \num{99.5}} & \num{50.0} \tiny{± \num{1.7}} & \num{65.6} \tiny{± \num{0.5}} & \num{4984} \tiny{± \num{65.0}} & \num{49.3} \tiny{± \num{1.1}} & \num{60.0} \tiny{± \num{8.5}} & \num{7266} \tiny{± \num{554.0}} & \num{39.3} \tiny{± \num{6.3}} & \num{43.3} \tiny{± \num{4.7}} & \num{6611} \tiny{± \num{473.3}} & \num{29.1} \tiny{± \num{4.6}} \\ 
\midrule
Laser-DE-L4096          & \makecell[r]{{\num{90.4} \tiny{± \num{0.1}}} \\ {\textcolor{ForestGreen}{\small{(+ \num{2.8})}}}} & \makecell[r]{\num{849} \tiny{± \num{1.3}} \\ \textcolor{BrickRed}{\small{(+ \num{328})}}} & \makecell[r]{\num{87.0} \tiny{± \num{0.2}} \\ \textcolor{BrickRed}{\small{(- \num{0.5})}}} & \makecell[r]{\num{61.9} \tiny{± \num{1.0}} \\ \textcolor{BrickRed}{\small{(- \num{1.7})}}} & \makecell[r]{\num{1767} \tiny{± \num{29.4}} \\ \textcolor{ForestGreen}{\small{(- \num{1750})}}} & \makecell[r]{\num{51.8} \tiny{± \num{1.0}} \\ \textcolor{ForestGreen}{\small{(+ \num{1.8})}}} & \makecell[r]{\num{71.1} \tiny{± \num{0.6}} \\ \textcolor{ForestGreen}{\small{(+ \num{5.5})}}} & \makecell[r]{\num{3358} \tiny{± \num{21.7}} \\ \textcolor{ForestGreen}{\small{(- \num{1626})}}} & \makecell[r]{\num{46.8} \tiny{± \num{0.6}} \\ \textcolor{BrickRed}{\small{(- \num{2.5})}}} & \makecell[r]{\num{66.7} \tiny{± \num{2.4}} \\ \textcolor{ForestGreen}{\small{(+ \num{6.7})}}} & \makecell[r]{\num{5563} \tiny{± \num{283.0}} \\ \textcolor{ForestGreen}{\small{(- \num{1703})}}} & \makecell[r]{\num{42.2} \tiny{± \num{2.2}} \\ \textcolor{ForestGreen}{\small{(+ \num{2.9})}}} & \makecell[r]{{\num{52.0} \tiny{± \num{5.6}}} \\ {\textcolor{ForestGreen}{\small{(+ \num{8.7})}}}} & \makecell[r]{\num{5553} \tiny{± \num{176.2}} \\ \textcolor{ForestGreen}{\small{(- \num{1058})}}} & \makecell[r]{{\num{35.0} \tiny{± \num{4.1}}} \\ {\textcolor{ForestGreen}{\small{(+ \num{5.9})}}}} \\ 
AdaptThink-delta0.05    & \makecell[r]{\num{88.8} \tiny{± \num{0.2}} \\ \textcolor{ForestGreen}{\small{(+ \num{1.2})}}} & \makecell[r]{\num{394} \tiny{± \num{2.0}} \\ \textcolor{ForestGreen}{\small{(- \num{127})}}} & \makecell[r]{\num{88.8} \tiny{± \num{0.2}} \\ \textcolor{ForestGreen}{\small{(+ \num{1.3})}}} & \makecell[r]{\num{62.0} \tiny{± \num{1.6}} \\ \textcolor{BrickRed}{\small{(- \num{1.6})}}} & \makecell[r]{\num{1983} \tiny{± \num{67.2}} \\ \textcolor{ForestGreen}{\small{(- \num{1534})}}} & \makecell[r]{\num{56.6} \tiny{± \num{2.1}} \\ \textcolor{ForestGreen}{\small{(+ \num{6.6})}}} & \makecell[r]{\num{67.2} \tiny{± \num{1.0}} \\ \textcolor{ForestGreen}{\small{(+ \num{1.6})}}} & \makecell[r]{\num{4080} \tiny{± \num{79.7}} \\ \textcolor{ForestGreen}{\small{(- \num{904})}}} & \makecell[r]{\num{56.4} \tiny{± \num{0.6}} \\ \textcolor{ForestGreen}{\small{(+ \num{7.1})}}} & \makecell[r]{\num{68.0} \tiny{± \num{3.8}} \\ \textcolor{ForestGreen}{\small{(+ \num{8.0})}}} & \makecell[r]{\num{6682} \tiny{± \num{762.7}} \\ \textcolor{ForestGreen}{\small{(- \num{584})}}} & \makecell[r]{\num{48.8} \tiny{± \num{2.2}} \\ \textcolor{ForestGreen}{\small{(+ \num{9.5})}}} & \makecell[r]{{\num{46.7} \tiny{± \num{2.4}}} \\ {\textcolor{ForestGreen}{\small{(+ \num{3.4})}}}} & \makecell[r]{\num{6599} \tiny{± \num{498.6}} \\ \textcolor{ForestGreen}{\small{(- \num{12})}}} & \makecell[r]{{\num{33.8} \tiny{± \num{2.3}}} \\ {\textcolor{ForestGreen}{\small{(+ \num{1.8})}}}} \\ 
\midrule
\textbf{When2Think (Ours)} & \makecell[r]{\num{89.3} \tiny{± \num{0.1}} \\ \textcolor{ForestGreen}{\small{(+ \num{1.7})}}} & \makecell[r]{\num{460} \tiny{± \num{2.3}} \\ \textcolor{ForestGreen}{\small{(- \num{61})}}} & \makecell[r]{\num{88.7} \tiny{± \num{0.1}} \\ \textcolor{ForestGreen}{\small{(+ \num{1.2})}}} & \makecell[r]{\num{63.2} \tiny{± \num{1.1}} \\ \textcolor{BrickRed}{\small{(- \num{0.4})}}} & \makecell[r]{\num{2451} \tiny{± \num{47.5}} \\ \textcolor{ForestGreen}{\small{(- \num{1066})}}} & \makecell[r]{\num{54.1} \tiny{± \num{0.8}} \\ \textcolor{ForestGreen}{\small{(+ \num{4.1})}}} & \makecell[r]{\num{69.1} \tiny{± \num{0.7}} \\ \textcolor{ForestGreen}{\small{(+ \num{3.5})}}} & \makecell[r]{\num{4250} \tiny{± \num{70.6}} \\ \textcolor{ForestGreen}{\small{(- \num{734})}}} & \makecell[r]{\num{55.4} \tiny{± \num{0.8}} \\ \textcolor{ForestGreen}{\small{(+ \num{6.1})}}} & \makecell[r]{\num{68.7} \tiny{± \num{4.5}} \\ \textcolor{ForestGreen}{\small{(+ \num{8.7})}}} & \makecell[r]{\num{7855} \tiny{± \num{397.9}} \\ \textcolor{BrickRed}{\small{(+ \num{589})}}} & \makecell[r]{\num{45.4} \tiny{± \num{3.7}} \\ \textcolor{ForestGreen}{\small{(+ \num{6.1})}}} & \makecell[r]{{\num{48.0} \tiny{± \num{1.8}}} \\ {\textcolor{ForestGreen}{\small{(+ \num{4.7})}}}} & \makecell[r]{\num{7347} \tiny{± \num{799.1}} \\ \textcolor{BrickRed}{\small{(+ \num{736})}}} & \makecell[r]{{\num{31.6} \tiny{± \num{1.4}}} \\ {\textcolor{ForestGreen}{\small{(+ \num{2.5})}}}} \\ 
\bottomrule
\bottomrule
\end{tabular}

        }
    \end{adjustbox}
\end{table}

\begin{table}[H]
    \centering
    \caption{\textbf{Performance comparison on MATH-500 stratified by problem difficulty (Levels~1--5).}
    The table reports \texttt{Pass@1} accuracy (\%), average tokens per response (↓ \texttt{Tokens}), and Accuracy--Cost Efficiency (↑ \texttt{ACE}) for each difficulty level under the same evaluation protocol as the main results. 
    Results are reported as mean~$\pm$~standard deviation over five runs. 
    Values in parentheses indicate differences relative to R1-Distill-Qwen (\textcolor{ForestGreen}{green}: improvement; \textcolor{BrickRed}{red}: degradation).}
    \label{tab:eval_math500}
    \begin{adjustbox}{angle=270, max height=\textheight}
        \resizebox{0.9\textheight}{!}
        {
            \begin{tabular}{l|rrr|rrr|rrr|rrr|rrr}
\toprule
\toprule
\multirowcell{3}{\diagbox{Model}{Benchmark}} & \multicolumn{13}{c}{\large{MATH-500}} \\ 
\cline{2-16} 
    & \multicolumn{3}{c|}{\large{Level 1}} & \multicolumn{3}{c|}{\large{Level 2}} & \multicolumn{3}{c|}{\large{Level 3}} & \multicolumn{3}{c|}{\large{Level 4}} & \multicolumn{3}{c}{\large{Level 5}} \\ 
\cline{2-16} 
    & \multicolumn{1}{c}{↑ Pass@1\small{(\%)}}
    & \multicolumn{1}{c}{↓ Tokens\small{(\#)}}
    & \multicolumn{1}{c|}{↑ ACE\small{(\%)}}
    & \multicolumn{1}{c}{↑ Pass@1\small{(\%)}}
    & \multicolumn{1}{c}{↓ Tokens\small{(\#)}}
    & \multicolumn{1}{c|}{↑ ACE\small{(\%)}}
    & \multicolumn{1}{c}{↑ Pass@1\small{(\%)}}
    & \multicolumn{1}{c}{↓ Tokens\small{(\#)}}
    & \multicolumn{1}{c|}{↑ ACE\small{(\%)}}
    & \multicolumn{1}{c}{↑ Pass@1\small{(\%)}}
    & \multicolumn{1}{c}{↓ Tokens\small{(\#)}}
    & \multicolumn{1}{c|}{↑ ACE\small{(\%)}}
    & \multicolumn{1}{c}{↑ Pass@1\small{(\%)}}
    & \multicolumn{1}{c}{↓ Tokens\small{(\#)}}
    & \multicolumn{1}{c}{↑ ACE\small{(\%)}}
 \\ 
\midrule
\midrule
\multicolumn{16}{l}{\large{\textit{\textbf{1.5B Models}}}} \\ 
\midrule
Qwen2.5-Instruct        & \num{79.5} \tiny{± \num{6.9}} & \num{342} \tiny{± \num{11.9}} & \multicolumn{1}{c|}{--} & \num{71.1} \tiny{± \num{4.2}} & \num{429} \tiny{± \num{15.8}} & \multicolumn{1}{c|}{--} & \num{57.0} \tiny{± \num{1.4}} & \num{768} \tiny{± \num{144.7}} & \multicolumn{1}{c|}{--} & \num{37.5} \tiny{± \num{4.0}} & \num{828} \tiny{± \num{213.7}} & \multicolumn{1}{c|}{--} & \num{21.3} \tiny{± \num{2.9}} & \num{1037} \tiny{± \num{113.9}} & \multicolumn{1}{c}{--} \\ 
Qwen2.5-Math-Instruct   & \num{94.0} \tiny{± \num{1.3}} & \num{346} \tiny{± \num{6.1}} & \multicolumn{1}{c|}{--} & \num{90.2} \tiny{± \num{0.5}} & \num{422} \tiny{± \num{4.4}} & \multicolumn{1}{c|}{--} & \num{87.8} \tiny{± \num{3.0}} & \num{499} \tiny{± \num{13.7}} & \multicolumn{1}{c|}{--} & \num{71.2} \tiny{± \num{2.3}} & \num{644} \tiny{± \num{26.0}} & \multicolumn{1}{c|}{--} & \num{51.8} \tiny{± \num{3.6}} & \num{774} \tiny{± \num{38.5}} & \multicolumn{1}{c}{--} \\ 
\midrule
\textbf{R1-Distill-Qwen (Base)} & \num{92.1} \tiny{± \num{2.7}} & \num{1199} \tiny{± \num{150.9}} & \num{89.3} \tiny{± \num{2.1}} & \num{91.8} \tiny{± \num{2.9}} & \num{1724} \tiny{± \num{136.2}} & \num{83.7} \tiny{± \num{2.9}} & \num{85.7} \tiny{± \num{2.7}} & \num{2530} \tiny{± \num{64.1}} & \num{74.7} \tiny{± \num{3.3}} & \num{79.4} \tiny{± \num{2.0}} & \num{4006} \tiny{± \num{197.3}} & \num{65.0} \tiny{± \num{3.6}} & \num{71.0} \tiny{± \num{2.3}} & \num{6830} \tiny{± \num{203.3}} & \num{56.6} \tiny{± \num{2.3}} \\ 
DeepScaleR-Preview      & \num{94.0} \tiny{± \num{2.1}} & \num{1250} \tiny{± \num{62.6}} & \num{86.3} \tiny{± \num{2.5}} & \num{95.8} \tiny{± \num{0.5}} & \num{1533} \tiny{± \num{48.4}} & \num{85.2} \tiny{± \num{0.6}} & \num{93.9} \tiny{± \num{1.4}} & \num{2135} \tiny{± \num{94.4}} & \num{78.3} \tiny{± \num{1.7}} & \num{87.7} \tiny{± \num{2.0}} & \num{2913} \tiny{± \num{138.3}} & \num{71.5} \tiny{± \num{1.0}} & \num{75.8} \tiny{± \num{2.1}} & \num{4703} \tiny{± \num{115.7}} & \num{62.3} \tiny{± \num{1.9}} \\ 
\midrule
LC-R1                   & \makecell[r]{{\num{90.2} \tiny{± \num{3.8}}} \\ {\textcolor{BrickRed}{\small{(- \num{1.9})}}}} & \makecell[r]{\num{626} \tiny{± \num{23.2}} \\ \textcolor{ForestGreen}{\small{(- \num{573})}}} & \makecell[r]{{\num{85.6} \tiny{± \num{3.5}}} \\ {\textcolor{BrickRed}{\small{(- \num{3.7})}}}} & \makecell[r]{{\num{85.8} \tiny{± \num{4.0}}} \\ {\textcolor{BrickRed}{\small{(- \num{6.0})}}}} & \makecell[r]{\num{916} \tiny{± \num{74.5}} \\ \textcolor{ForestGreen}{\small{(- \num{808})}}} & \makecell[r]{{\num{78.9} \tiny{± \num{3.4}}} \\ {\textcolor{BrickRed}{\small{(- \num{4.8})}}}} & \makecell[r]{{\num{81.9} \tiny{± \num{1.8}}} \\ {\textcolor{BrickRed}{\small{(- \num{3.8})}}}} & \makecell[r]{\num{1491} \tiny{± \num{116.7}} \\ \textcolor{ForestGreen}{\small{(- \num{1039})}}} & \makecell[r]{{\num{71.0} \tiny{± \num{2.5}}} \\ {\textcolor{BrickRed}{\small{(- \num{3.7})}}}} & \makecell[r]{{\num{77.3} \tiny{± \num{3.5}}} \\ {\textcolor{BrickRed}{\small{(- \num{2.1})}}}} & \makecell[r]{\num{2265} \tiny{± \num{202.1}} \\ \textcolor{ForestGreen}{\small{(- \num{1741})}}} & \makecell[r]{{\num{66.0} \tiny{± \num{4.2}}} \\ {\textcolor{ForestGreen}{\small{(+ \num{1.0})}}}} & \makecell[r]{{\num{59.3} \tiny{± \num{2.4}}} \\ {\textcolor{BrickRed}{\small{(- \num{11.7})}}}} & \makecell[r]{\num{3880} \tiny{± \num{313.9}} \\ \textcolor{ForestGreen}{\small{(- \num{2950})}}} & \makecell[r]{{\num{47.3} \tiny{± \num{2.0}}} \\ {\textcolor{BrickRed}{\small{(- \num{9.3})}}}} \\ 
ThinkPrune-iter3k       & \makecell[r]{{\num{93.0} \tiny{± \num{5.5}}} \\ {\textcolor{ForestGreen}{\small{(+ \num{0.9})}}}} & \makecell[r]{\num{770} \tiny{± \num{20.3}} \\ \textcolor{ForestGreen}{\small{(- \num{429})}}} & \makecell[r]{{\num{80.9} \tiny{± \num{5.5}}} \\ {\textcolor{BrickRed}{\small{(- \num{8.9})}}}} & \makecell[r]{{\num{91.3} \tiny{± \num{2.9}}} \\ {\textcolor{BrickRed}{\small{(- \num{0.5})}}}} & \makecell[r]{\num{990} \tiny{± \num{27.3}} \\ \textcolor{ForestGreen}{\small{(- \num{734})}}} & \makecell[r]{{\num{72.8} \tiny{± \num{4.0}}} \\ {\textcolor{BrickRed}{\small{(- \num{10.9})}}}} & \makecell[r]{{\num{90.3} \tiny{± \num{3.4}}} \\ {\textcolor{ForestGreen}{\small{(+ \num{4.6})}}}} & \makecell[r]{\num{1421} \tiny{± \num{68.5}} \\ \textcolor{ForestGreen}{\small{(- \num{1109})}}} & \makecell[r]{{\num{67.9} \tiny{± \num{3.4}}} \\ {\textcolor{BrickRed}{\small{(- \num{6.8})}}}} & \makecell[r]{{\num{83.3} \tiny{± \num{4.0}}} \\ {\textcolor{ForestGreen}{\small{(+ \num{3.9})}}}} & \makecell[r]{\num{2106} \tiny{± \num{138.9}} \\ \textcolor{ForestGreen}{\small{(- \num{1900})}}} & \makecell[r]{{\num{60.9} \tiny{± \num{3.0}}} \\ {\textcolor{BrickRed}{\small{(- \num{4.1})}}}} & \makecell[r]{{\num{66.6} \tiny{± \num{4.3}}} \\ {\textcolor{BrickRed}{\small{(- \num{4.4})}}}} & \makecell[r]{\num{3068} \tiny{± \num{68.4}} \\ \textcolor{ForestGreen}{\small{(- \num{3762})}}} & \makecell[r]{{\num{48.5} \tiny{± \num{3.1}}} \\ {\textcolor{BrickRed}{\small{(- \num{8.1})}}}} \\ 
Laser-DE-L4096          & \makecell[r]{{\num{91.2} \tiny{± \num{1.9}}} \\ {\textcolor{BrickRed}{\small{(- \num{0.9})}}}} & \makecell[r]{\num{898} \tiny{± \num{68.1}} \\ \textcolor{ForestGreen}{\small{(- \num{301})}}} & \makecell[r]{\num{82.7} \tiny{± \num{2.7}} \\ \textcolor{BrickRed}{\small{(- \num{6.6})}}} & \makecell[r]{\num{90.2} \tiny{± \num{1.8}} \\ \textcolor{BrickRed}{\small{(- \num{1.6})}}} & \makecell[r]{\num{1211} \tiny{± \num{23.1}} \\ \textcolor{ForestGreen}{\small{(- \num{513})}}} & \makecell[r]{\num{78.1} \tiny{± \num{2.2}} \\ \textcolor{BrickRed}{\small{(- \num{5.6})}}} & \makecell[r]{\num{92.2} \tiny{± \num{2.1}} \\ \textcolor{ForestGreen}{\small{(+ \num{6.5})}}} & \makecell[r]{\num{1763} \tiny{± \num{59.8}} \\ \textcolor{ForestGreen}{\small{(- \num{767})}}} & \makecell[r]{\num{72.0} \tiny{± \num{2.6}} \\ \textcolor{BrickRed}{\small{(- \num{2.7})}}} & \makecell[r]{\num{81.6} \tiny{± \num{0.4}} \\ \textcolor{ForestGreen}{\small{(+ \num{2.2})}}} & \makecell[r]{\num{2551} \tiny{± \num{113.8}} \\ \textcolor{ForestGreen}{\small{(- \num{1455})}}} & \makecell[r]{\num{63.0} \tiny{± \num{1.4}} \\ \textcolor{BrickRed}{\small{(- \num{2.0})}}} & \makecell[r]{{\num{69.6} \tiny{± \num{3.3}}} \\ {\textcolor{BrickRed}{\small{(- \num{1.4})}}}} & \makecell[r]{\num{4056} \tiny{± \num{90.1}} \\ \textcolor{ForestGreen}{\small{(- \num{2774})}}} & \makecell[r]{{\num{53.7} \tiny{± \num{2.3}}} \\ {\textcolor{BrickRed}{\small{(- \num{2.9})}}}} \\ 
AdaptThink-delta0.05    & \makecell[r]{\num{94.4} \tiny{± \num{2.1}} \\ \textcolor{ForestGreen}{\small{(+ \num{2.3})}}} & \makecell[r]{\num{512} \tiny{± \num{156.9}} \\ \textcolor{ForestGreen}{\small{(- \num{687})}}} & \makecell[r]{\num{94.4} \tiny{± \num{2.1}} \\ \textcolor{ForestGreen}{\small{(+ \num{5.1})}}} & \makecell[r]{\num{92.0} \tiny{± \num{1.4}} \\ \textcolor{ForestGreen}{\small{(+ \num{0.2})}}} & \makecell[r]{\num{510} \tiny{± \num{29.7}} \\ \textcolor{ForestGreen}{\small{(- \num{1214})}}} & \makecell[r]{\num{91.8} \tiny{± \num{1.6}} \\ \textcolor{ForestGreen}{\small{(+ \num{8.1})}}} & \makecell[r]{\num{89.3} \tiny{± \num{1.2}} \\ \textcolor{ForestGreen}{\small{(+ \num{3.6})}}} & \makecell[r]{\num{1029} \tiny{± \num{87.5}} \\ \textcolor{ForestGreen}{\small{(- \num{1501})}}} & \makecell[r]{\num{88.0} \tiny{± \num{1.1}} \\ \textcolor{ForestGreen}{\small{(+ \num{13.3})}}} & \makecell[r]{\num{80.0} \tiny{± \num{3.2}} \\ \textcolor{BrickRed}{\small{(- \num{0.6})}}} & \makecell[r]{\num{1477} \tiny{± \num{63.3}} \\ \textcolor{ForestGreen}{\small{(- \num{2529})}}} & \makecell[r]{\num{79.2} \tiny{± \num{3.2}} \\ \textcolor{ForestGreen}{\small{(+ \num{14.2})}}} & \makecell[r]{{\num{65.2} \tiny{± \num{3.2}}} \\ {\textcolor{BrickRed}{\small{(- \num{5.8})}}}} & \makecell[r]{\num{2525} \tiny{± \num{112.2}} \\ \textcolor{ForestGreen}{\small{(- \num{4305})}}} & \makecell[r]{{\num{63.9} \tiny{± \num{2.7}}} \\ {\textcolor{ForestGreen}{\small{(+ \num{7.3})}}}} \\ 
Thinkless-RL            & \makecell[r]{\num{95.8} \tiny{± \num{3.4}} \\ \textcolor{ForestGreen}{\small{(+ \num{3.7})}}} & \makecell[r]{\num{1374} \tiny{± \num{196.8}} \\ \textcolor{BrickRed}{\small{(+ \num{175})}}} & \makecell[r]{\num{90.8} \tiny{± \num{4.1}} \\ \textcolor{ForestGreen}{\small{(+ \num{1.5})}}} & \makecell[r]{\num{94.7} \tiny{± \num{2.7}} \\ \textcolor{ForestGreen}{\small{(+ \num{2.9})}}} & \makecell[r]{\num{1867} \tiny{± \num{146.3}} \\ \textcolor{BrickRed}{\small{(+ \num{143})}}} & \makecell[r]{\num{89.2} \tiny{± \num{2.3}} \\ \textcolor{ForestGreen}{\small{(+ \num{5.5})}}} & \makecell[r]{\num{91.6} \tiny{± \num{2.1}} \\ \textcolor{ForestGreen}{\small{(+ \num{5.9})}}} & \makecell[r]{\num{2669} \tiny{± \num{218.2}} \\ \textcolor{BrickRed}{\small{(+ \num{139})}}} & \makecell[r]{\num{82.3} \tiny{± \num{3.1}} \\ \textcolor{ForestGreen}{\small{(+ \num{7.6})}}} & \makecell[r]{\num{82.2} \tiny{± \num{3.2}} \\ \textcolor{ForestGreen}{\small{(+ \num{2.8})}}} & \makecell[r]{\num{3837} \tiny{± \num{221.7}} \\ \textcolor{ForestGreen}{\small{(- \num{169})}}} & \makecell[r]{\num{71.2} \tiny{± \num{1.8}} \\ \textcolor{ForestGreen}{\small{(+ \num{6.2})}}} & \makecell[r]{{\num{71.3} \tiny{± \num{3.4}}} \\ {\textcolor{ForestGreen}{\small{(+ \num{0.3})}}}} & \makecell[r]{\num{5668} \tiny{± \num{145.3}} \\ \textcolor{ForestGreen}{\small{(- \num{1162})}}} & \makecell[r]{{\num{58.2} \tiny{± \num{3.1}}} \\ {\textcolor{ForestGreen}{\small{(+ \num{1.6})}}}} \\ 
\midrule
\textbf{When2Think (Ours)} & \makecell[r]{\num{95.8} \tiny{± \num{1.9}} \\ \textcolor{ForestGreen}{\small{(+ \num{3.7})}}} & \makecell[r]{\num{619} \tiny{± \num{11.0}} \\ \textcolor{ForestGreen}{\small{(- \num{580})}}} & \makecell[r]{\num{94.3} \tiny{± \num{1.9}} \\ \textcolor{ForestGreen}{\small{(+ \num{5.0})}}} & \makecell[r]{\num{92.9} \tiny{± \num{2.6}} \\ \textcolor{ForestGreen}{\small{(+ \num{1.1})}}} & \makecell[r]{\num{862} \tiny{± \num{76.1}} \\ \textcolor{ForestGreen}{\small{(- \num{862})}}} & \makecell[r]{\num{89.2} \tiny{± \num{2.4}} \\ \textcolor{ForestGreen}{\small{(+ \num{5.5})}}} & \makecell[r]{\num{90.5} \tiny{± \num{0.7}} \\ \textcolor{ForestGreen}{\small{(+ \num{4.8})}}} & \makecell[r]{\num{1447} \tiny{± \num{42.6}} \\ \textcolor{ForestGreen}{\small{(- \num{1083})}}} & \makecell[r]{\num{82.8} \tiny{± \num{0.6}} \\ \textcolor{ForestGreen}{\small{(+ \num{8.1})}}} & \makecell[r]{\num{84.5} \tiny{± \num{2.2}} \\ \textcolor{ForestGreen}{\small{(+ \num{5.1})}}} & \makecell[r]{\num{2277} \tiny{± \num{157.7}} \\ \textcolor{ForestGreen}{\small{(- \num{1729})}}} & \makecell[r]{\num{76.6} \tiny{± \num{1.9}} \\ \textcolor{ForestGreen}{\small{(+ \num{11.6})}}} & \makecell[r]{{\num{70.7} \tiny{± \num{2.9}}} \\ {\textcolor{BrickRed}{\small{(- \num{0.3})}}}} & \makecell[r]{\num{4554} \tiny{± \num{177.5}} \\ \textcolor{ForestGreen}{\small{(- \num{2276})}}} & \makecell[r]{{\num{60.3} \tiny{± \num{2.0}}} \\ {\textcolor{ForestGreen}{\small{(+ \num{3.7})}}}} \\ 
\midrule
\multicolumn{16}{l}{\large{\textit{\textbf{7B Models}}}} \\ 
\midrule
Qwen2.5-Instruct        & \num{96.3} \tiny{± \num{1.3}} & \num{342} \tiny{± \num{4.9}} & \multicolumn{1}{c|}{--} & \num{92.4} \tiny{± \num{1.4}} & \num{440} \tiny{± \num{4.2}} & \multicolumn{1}{c|}{--} & \num{87.4} \tiny{± \num{2.8}} & \num{506} \tiny{± \num{7.4}} & \multicolumn{1}{c|}{--} & \num{72.7} \tiny{± \num{3.0}} & \num{623} \tiny{± \num{8.6}} & \multicolumn{1}{c|}{--} & \num{53.9} \tiny{± \num{3.9}} & \num{775} \tiny{± \num{7.1}} & \multicolumn{1}{c}{--} \\ 
Qwen2.5-Math-Instruct   & \num{96.7} \tiny{± \num{2.7}} & \num{370} \tiny{± \num{16.9}} & \multicolumn{1}{c|}{--} & \num{93.8} \tiny{± \num{1.0}} & \num{449} \tiny{± \num{7.7}} & \multicolumn{1}{c|}{--} & \num{93.9} \tiny{± \num{1.1}} & \num{571} \tiny{± \num{18.8}} & \multicolumn{1}{c|}{--} & \num{81.6} \tiny{± \num{0.9}} & \num{767} \tiny{± \num{13.6}} & \multicolumn{1}{c|}{--} & \num{64.6} \tiny{± \num{2.7}} & \num{1013} \tiny{± \num{44.0}} & \multicolumn{1}{c}{--} \\ 
\midrule
R1-Distill-Qwen & \num{93.5} \tiny{± \num{1.0}} & \num{948} \tiny{± \num{83.4}} & \num{90.1} \tiny{± \num{2.2}} & \num{93.3} \tiny{± \num{1.1}} & \num{1306} \tiny{± \num{33.5}} & \num{86.5} \tiny{± \num{1.2}} & \num{95.6} \tiny{± \num{1.6}} & \num{1815} \tiny{± \num{50.9}} & \num{84.1} \tiny{± \num{2.3}} & \num{90.5} \tiny{± \num{1.7}} & \num{2800} \tiny{± \num{24.8}} & \num{77.7} \tiny{± \num{1.4}} & \num{84.8} \tiny{± \num{1.5}} & \num{4449} \tiny{± \num{163.9}} & \num{71.7} \tiny{± \num{2.0}} \\ 
\midrule
Laser-DE-L4096          & \makecell[r]{{\num{96.7} \tiny{± \num{1.3}}} \\ {\textcolor{ForestGreen}{\small{(+ \num{3.2})}}}} & \makecell[r]{\num{953} \tiny{± \num{10.9}} \\ \textcolor{BrickRed}{\small{(+ \num{5})}}} & \makecell[r]{\num{82.9} \tiny{± \num{1.5}} \\ \textcolor{BrickRed}{\small{(- \num{7.2})}}} & \makecell[r]{\num{97.3} \tiny{± \num{1.3}} \\ \textcolor{ForestGreen}{\small{(+ \num{4.0})}}} & \makecell[r]{\num{1080} \tiny{± \num{21.8}} \\ \textcolor{ForestGreen}{\small{(- \num{226})}}} & \makecell[r]{\num{78.4} \tiny{± \num{0.9}} \\ \textcolor{BrickRed}{\small{(- \num{8.1})}}} & \makecell[r]{\num{98.9} \tiny{± \num{1.0}} \\ \textcolor{ForestGreen}{\small{(+ \num{3.3})}}} & \makecell[r]{\num{1326} \tiny{± \num{32.3}} \\ \textcolor{ForestGreen}{\small{(- \num{489})}}} & \makecell[r]{\num{75.7} \tiny{± \num{1.5}} \\ \textcolor{BrickRed}{\small{(- \num{10.4})}}} & \makecell[r]{\num{94.4} \tiny{± \num{1.0}} \\ \textcolor{ForestGreen}{\small{(+ \num{3.9})}}} & \makecell[r]{\num{1782} \tiny{± \num{41.2}} \\ \textcolor{ForestGreen}{\small{(- \num{1018})}}} & \makecell[r]{\num{71.7} \tiny{± \num{0.6}} \\ \textcolor{BrickRed}{\small{(- \num{6.0})}}} & \makecell[r]{{\num{86.9} \tiny{± \num{1.1}}} \\ {\textcolor{ForestGreen}{\small{(+ \num{2.1})}}}} & \makecell[r]{\num{2717} \tiny{± \num{23.4}} \\ \textcolor{ForestGreen}{\small{(- \num{1732})}}} & \makecell[r]{{\num{66.3} \tiny{± \num{1.1}}} \\ {\textcolor{BrickRed}{\small{(- \num{5.4})}}}} \\ 
AdaptThink-delta0.05    & \makecell[r]{\num{95.3} \tiny{± \num{1.6}} \\ \textcolor{ForestGreen}{\small{(+ \num{1.8})}}} & \makecell[r]{\num{420} \tiny{± \num{12.7}} \\ \textcolor{ForestGreen}{\small{(- \num{528})}}} & \makecell[r]{\num{95.2} \tiny{± \num{1.7}} \\ \textcolor{ForestGreen}{\small{(+ \num{5.1})}}} & \makecell[r]{\num{95.6} \tiny{± \num{1.8}} \\ \textcolor{ForestGreen}{\small{(+ \num{2.3})}}} & \makecell[r]{\num{568} \tiny{± \num{16.8}} \\ \textcolor{ForestGreen}{\small{(- \num{738})}}} & \makecell[r]{\num{95.1} \tiny{± \num{1.7}} \\ \textcolor{ForestGreen}{\small{(+ \num{8.6})}}} & \makecell[r]{\num{97.5} \tiny{± \num{1.1}} \\ \textcolor{ForestGreen}{\small{(+ \num{1.9})}}} & \makecell[r]{\num{929} \tiny{± \num{24.6}} \\ \textcolor{ForestGreen}{\small{(- \num{886})}}} & \makecell[r]{\num{95.2} \tiny{± \num{1.8}} \\ \textcolor{ForestGreen}{\small{(+ \num{11.1})}}} & \makecell[r]{\num{92.0} \tiny{± \num{1.7}} \\ \textcolor{ForestGreen}{\small{(+ \num{1.5})}}} & \makecell[r]{\num{1416} \tiny{± \num{63.8}} \\ \textcolor{ForestGreen}{\small{(- \num{1384})}}} & \makecell[r]{\num{89.0} \tiny{± \num{1.6}} \\ \textcolor{ForestGreen}{\small{(+ \num{11.3})}}} & \makecell[r]{{\num{85.2} \tiny{± \num{1.8}}} \\ {\textcolor{ForestGreen}{\small{(+ \num{0.4})}}}} & \makecell[r]{\num{3186} \tiny{± \num{171.3}} \\ \textcolor{ForestGreen}{\small{(- \num{1263})}}} & \makecell[r]{{\num{78.8} \tiny{± \num{1.8}}} \\ {\textcolor{ForestGreen}{\small{(+ \num{7.1})}}}} \\ 
\midrule
\textbf{When2Think (Ours)} & \makecell[r]{\num{96.7} \tiny{± \num{2.1}} \\ \textcolor{ForestGreen}{\small{(+ \num{3.2})}}} & \makecell[r]{\num{603} \tiny{± \num{23.9}} \\ \textcolor{ForestGreen}{\small{(- \num{345})}}} & \makecell[r]{\num{95.7} \tiny{± \num{2.3}} \\ \textcolor{ForestGreen}{\small{(+ \num{5.6})}}} & \makecell[r]{\num{98.7} \tiny{± \num{0.9}} \\ \textcolor{ForestGreen}{\small{(+ \num{5.4})}}} & \makecell[r]{\num{904} \tiny{± \num{33.8}} \\ \textcolor{ForestGreen}{\small{(- \num{402})}}} & \makecell[r]{\num{93.1} \tiny{± \num{1.2}} \\ \textcolor{ForestGreen}{\small{(+ \num{6.6})}}} & \makecell[r]{\num{97.0} \tiny{± \num{1.4}} \\ \textcolor{ForestGreen}{\small{(+ \num{1.4})}}} & \makecell[r]{\num{1247} \tiny{± \num{67.0}} \\ \textcolor{ForestGreen}{\small{(- \num{568})}}} & \makecell[r]{\num{91.4} \tiny{± \num{1.8}} \\ \textcolor{ForestGreen}{\small{(+ \num{7.3})}}} & \makecell[r]{\num{93.4} \tiny{± \num{1.2}} \\ \textcolor{ForestGreen}{\small{(+ \num{2.9})}}} & \makecell[r]{\num{2082} \tiny{± \num{58.8}} \\ \textcolor{ForestGreen}{\small{(- \num{718})}}} & \makecell[r]{\num{83.2} \tiny{± \num{1.0}} \\ \textcolor{ForestGreen}{\small{(+ \num{5.5})}}} & \makecell[r]{{\num{87.6} \tiny{± \num{2.0}}} \\ {\textcolor{ForestGreen}{\small{(+ \num{2.8})}}}} & \makecell[r]{\num{3825} \tiny{± \num{101.6}} \\ \textcolor{ForestGreen}{\small{(- \num{624})}}} & \makecell[r]{{\num{76.9} \tiny{± \num{1.4}}} \\ {\textcolor{ForestGreen}{\small{(+ \num{5.2})}}}} \\ 
\bottomrule
\bottomrule
\end{tabular}

        }
    \end{adjustbox}
\end{table}

\clearpage

\subsection{Cross-Domain Transfer on MMLU-Pro Stratified}
\label{appx:domain}

MMLU-Pro~\citep{NEURIPS2024_ad236edc} extends MMLU~\citep{hendrycks2021measuring} by filtering trivial or noisy questions, incorporating more reasoning-intensive problems, and expanding the answer set from four to typically ten options. We evaluate on MMLU-Pro Stratified~\citep{shi-etal-2025-educationq}, a balanced \num{1300}-question subset spanning 13 disciplines and 10 empirical difficulty levels. Each discipline--difficulty combination contains ten questions, and the difficulty levels are derived from mean item-level accuracy across ten strong models rather than from native MMLU-Pro annotations. We aggregate the levels into Easy (Levels 1--3), Medium (Levels 4--7), and Hard (Levels 8--10). The 13 disciplines are grouped into four supercategories: STEM (Biology, Chemistry, Computer Science, Engineering, Mathematics, and Physics), Humanities (History and Philosophy), Social Science (Economics and Psychology), and Professional disciplines (Business, Health, and Law).

\begin{figure}[H]
    \centering
    \includegraphics[width=\textwidth]{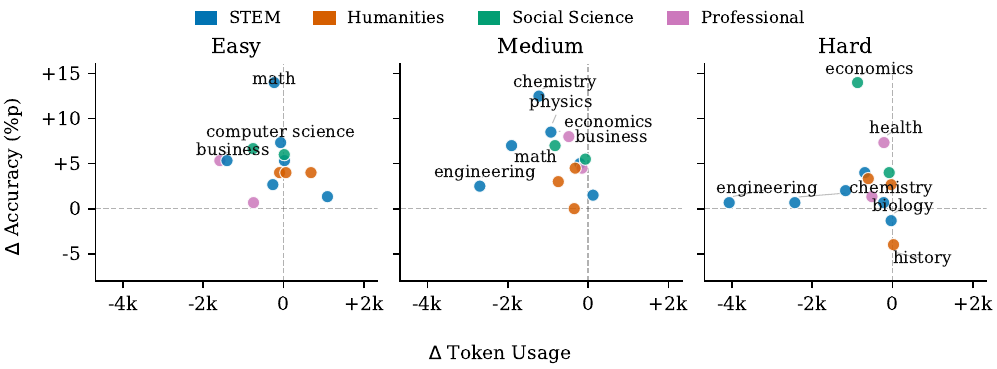}
    \caption{\textbf{Discipline-level accuracy and computation changes relative to R1-Distill on MMLU-Pro Stratified.} Each point represents one discipline, colored by supercategory, and the three panels correspond to the Easy, Medium, and Hard groups. The horizontal axis shows the change in average token usage, and the vertical axis shows the change in Pass@3 accuracy. The upper-left quadrant indicates simultaneous accuracy improvement and computation reduction. Accuracy gains and token reductions exhibit heterogeneous patterns across disciplines and difficulty groups.}
    \label{fig:mmlu_supercategory_transfer}
\end{figure}

\paragraph{Cross-domain transfer.}
Relative to R1-Distill, When2Think improves macro-average accuracy by 5.1, 5.3, and 2.7 percentage points on the Easy, Medium, and Hard groups, respectively (\cref{tab:mmlu_macro_transfer}). Average token usage decreases by 255, 752, and 842 tokens, corresponding to relative reductions of 8.1\%, 14.8\%, and 16.0\%. The corresponding changes in \textsc{Think} ratio are -4.0, -1.3, and -0.4 percentage points. At the discipline level, When2Think improves Easy-set accuracy in all 13 disciplines and simultaneously improves accuracy and reduces token usage on the Hard set in 11 of 13 disciplines (\cref{fig:mmlu_supercategory_transfer}). The form of transfer nevertheless varies across disciplines. On the Hard set, Economics exhibits the largest accuracy gain at +\num{14.0} percentage points, whereas the largest absolute token reductions occur in Engineering (-\num{4070}), Chemistry (-\num{2440}), Mathematics (-\num{1170}), and Computer Science (-\num{690}). Token reduction is limited in Physics, while History shows no simultaneous Hard-set improvement in accuracy and computation. Thus, accuracy gains and computation reductions do not follow the same disciplinary distribution.

\paragraph{Difficulty-conditioned behavior.}
The absolute behavior of When2Think also varies with empirical difficulty (\cref{tab:mmlu_discipline_behavior}). Macro-average Pass@3 decreases from 54.9\% on Easy questions to 41.6\% and 29.6\% on Medium and Hard questions, respectively. Over the same groups, average token usage increases from \num{2891} to \num{4343} and \num{4421} tokens, while the \textsc{Think} ratio increases from 94.6\% to 96.4\% and 97.1\%. Thus, increasing difficulty is accompanied by greater computation and more frequent use of explicit reasoning.

\clearpage

\paragraph{Mode selection versus computation allocation.}
Reasoning-mode selection and computation allocation exhibit distinct cross-domain patterns. Direct answering transfers most strongly to Mathematics, where the \textsc{Think} ratio is 64.0\%, 64.0\%, and 70.0\% on Easy, Medium, and Hard questions, respectively (\cref{tab:mmlu_discipline_behavior}). Despite identical \textsc{Think} ratios on Easy and Medium questions, average token usage increases from \num{1721} to \num{3602} tokens, showing that mode frequency alone does not determine computation allocation. Outside Mathematics, explicit reasoning remains predominant, with non-Math average \textsc{Think} ratios of 97.2\%, 99.0\%, and 99.3\% across the three groups. Several disciplines, including Economics, Health, History, Law, and Philosophy, retain a 100\% \textsc{Think} ratio while exhibiting substantial variation in token usage.

The baseline-relative comparison reinforces this distinction. On Hard questions, the macro-average \textsc{Think} ratio changes by only $-0.4$ percentage points, while average token usage decreases by \num{842} tokens (\cref{tab:mmlu_macro_transfer}). The Hard-set computation reduction therefore is not primarily accompanied by increased \textsc{NoThink} selection. The large token reductions in Engineering and Chemistry likewise occur despite \textsc{Think} ratios at or near 100\% (\cref{fig:mmlu_supercategory_transfer}, \cref{tab:mmlu_discipline_behavior}). Together, these observations are consistent with domain-selective transfer of direct answering and broader transfer of computation-allocation behavior. Because IDAC, BWS, and hybrid exploration are trained jointly, this interpretation constitutes behavioral attribution rather than a controlled causal decomposition of individual components.

\begin{table}[H] 
\centering \small \setlength{\tabcolsep}{5pt} 
\caption{\textbf{Macro-average cross-domain transfer relative to R1-Distill on MMLU-Pro Stratified.}
Accuracy and \textsc{Think}-ratio changes are reported in percentage points,
while token changes are reported in tokens per response.
Relative token changes are computed with respect to R1-Distill.}
\label{tab:mmlu_macro_transfer} 
\begin{tabular}{lrrr} 
\toprule
\toprule 
Metric & Easy & Medium & Hard \\ 
\midrule 
$\Delta$ Accuracy (\%p) & $+5.1$ & $+5.3$ & $+2.7$ \\ 
$\Delta$ Tokens & $-255$ & $-752$ & $-842$ \\ 
Relative token change & $-8.1\%$ & $-14.8\%$ & $-16.0\%$ \\ 
$\Delta$ \textsc{Think} ratio (\%p) & $-4.0$ & $-1.3$ & $-0.4$ \\ 
\bottomrule 
\bottomrule
\end{tabular} 
\end{table}

\begin{table}[H] 
\centering \scriptsize \setlength{\tabcolsep}{3.0pt} 
\caption{\textbf{Discipline-level behavior of When2Think on MMLU-Pro Stratified.}
Disciplines are grouped into four supercategories.
Pass@3 accuracy, average token usage, and \textsc{Think} ratio are reported
for the Easy, Medium, and Hard groups.
Macro Avg. includes all 13 disciplines, while Non-Math Avg. excludes Mathematics.}
\label{tab:mmlu_discipline_behavior} 
\resizebox{\textwidth}{!}{ 
\begin{tabular}{llrrrrrrrrr} 
\toprule
\toprule
& & \multicolumn{3}{c}{Pass@3 (\%)} & \multicolumn{3}{c}{Tokens} & \multicolumn{3}{c}{\textsc{Think} ratio (\%)} \\ \cmidrule(lr){3-5} \cmidrule(lr){6-8} \cmidrule(lr){9-11} Supercategory & Discipline & Easy & Medium & Hard & Easy & Medium & Hard & Easy & Medium & Hard \\ 
\midrule 
\multirow{6}{*}{STEM} & Biology & 41.3 & 35.0 & 36.0 & 3269 & 4819 & 4267 & 99.3 & 99.5 & 99.3 \\ & Chemistry & 71.3 & 51.0 & 35.3 & 4741 & 8755 & 7496 & 94.0 & 97.0 & 96.7 \\ & Computer Science & 75.3 & 40.0 & 32.0 & 2472 & 4782 & 4851 & 88.0 & 96.0 & 100.0 \\ & Engineering & 65.3 & 35.5 & 22.7 & 6656 & 10360 & 10765 & 97.3 & 99.5 & 100.0 \\ & \textbf{Mathematics} & \textbf{90.0} & \textbf{79.5} & \textbf{37.3} & \textbf{1721} & \textbf{3602} & \textbf{5379} & \textbf{64.0} & \textbf{64.0} & \textbf{70.0} \\ & Physics & 72.7 & 58.0 & 36.0 & 5775 & 6870 & 7872 & 92.7 & 99.5 & 99.3 \\ 
\midrule 
\multirow{2}{*}{Humanities} & History & 31.3 & 24.0 & 20.0 & 2009 & 1298 & 2202 & 100.0 & 100.0 & 100.0 \\ & Philosophy & 39.3 & 33.5 & 21.3 & 1425 & 2610 & 1764 & 100.0 & 100.0 & 100.0 \\ 
\midrule 
\multirow{2}{*}{Social Science} & Economics & 59.3 & 39.0 & 38.0 & 2458 & 1972 & 1913 & 100.0 & 100.0 & 100.0 \\ & Psychology & 41.3 & 39.5 & 29.3 & 820 & 1446 & 1338 & 100.0 & 98.5 & 100.0 \\ 
\midrule 
\multirow{3}{*}{Professional} & Business & 57.3 & 44.0 & 24.0 & 3205 & 6905 & 6084 & 94.7 & 98.5 & 96.7 \\ & Health & 36.0 & 33.5 & 30.0 & 1279 & 1526 & 1719 & 100.0 & 100.0 & 100.0 \\ & Law & 32.7 & 28.0 & 23.3 & 1752 & 1512 & 1827 & 100.0 & 100.0 & 100.0 \\ 
\midrule 
\multicolumn{2}{l}{\textbf{Macro Avg.}} & \textbf{54.9} & \textbf{41.6} & \textbf{29.6} & \textbf{2891} & \textbf{4343} & \textbf{4421} & \textbf{94.6} & \textbf{96.4} & \textbf{97.1} \\ \multicolumn{2}{l}{\textbf{Non-Math Avg.}} & \textbf{51.9} & \textbf{38.4} & \textbf{29.0} & \textbf{2988} & \textbf{4405} & \textbf{4342} & \textbf{97.2} & \textbf{99.0} & \textbf{99.3} \\ \bottomrule 
\bottomrule
\end{tabular} 
} 
\end{table}

\clearpage

\section{Ablation Studies}
\label{appx:ablation}

\subsection{Reward and Advantage Landscapes under IDAC and BWS}
\label{app:reward_advantage_landscape}

\paragraph{BWS versus GRPO in the Advantage Landscape.}
We analyze why Batch-Wise Standardization (BWS) is critical for optimizing IDAC by comparing three reward formulations: the raw \textsc{Think} reward, BWS advantages, and GRPO-style group-relative advantages.
We construct a Monte Carlo simulation using offline statistics from R1-Distill-Qwen-1.5B on DeepScaleR. For each problem, empirical accuracy $\alpha$ captures difficulty and reference length $\tau$ approximates typical reasoning cost. We then generate \textsc{Think} trajectories by perturbing $\tau$ with log-normal noise and sampling correctness according to $\alpha$.
This setup isolates how different normalization schemes allocate credit to reasoning trajectories as a function of problem difficulty and token usage.

\begin{figure}[H]
    \centering
    \includegraphics[width=\textwidth]{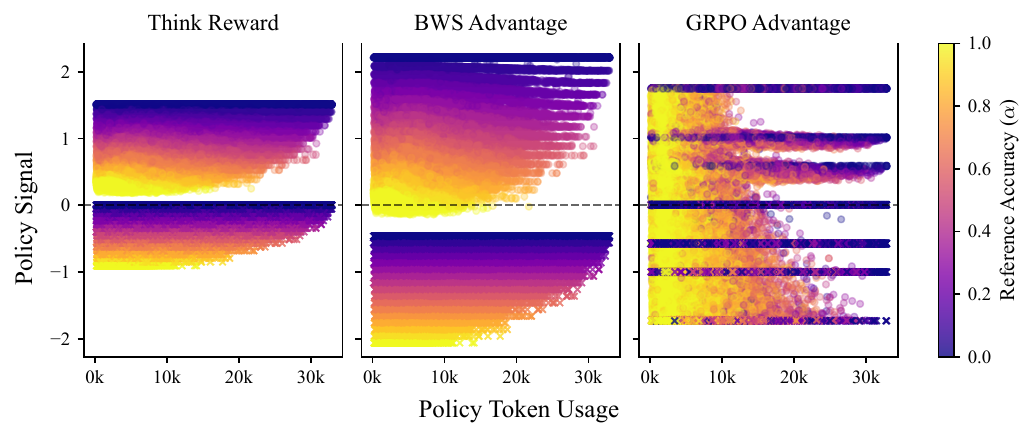}
    \caption{\textbf{Empirically Grounded Advantage Landscape of Explicit Reasoning.}
    Policy signals for simulated \textsc{Think} trajectories constructed from R1-Distill-Qwen-1.5B statistics on DeepScaleR. Colors represent reference accuracy $\alpha$ (hard problems in \textcolor{Blue}{blue}, $\alpha \to 0$, to easy problems in \textcolor{YellowOrange}{yellow}, $\alpha \to 1$), and markers indicate correctness (circles for correct, crosses for incorrect).
    \textbf{Left:} Raw IDAC \textsc{Think} reward with absolute length-sensitive bonus.
    \textbf{Middle:} BWS advantage (ours), which preserves positive advantages for long successful trajectories on hard problems.
    \textbf{Right:} GRPO-style advantage, where problem instance normalization yields fragmented signals and horizontal banding.}
    \label{fig:advantage}
\end{figure}

\paragraph{BWS: Preserving Global Difficulty Signals.}
BWS converts raw IDAC rewards into relative advantages computed over the full mini-batch, rather than normalizing within each instance. This batch-wise formulation preserves cross-instance difficulty structure that is lost under instance-local normalization.
As a result, trajectories that solve difficult problems can remain above the batch expectation and receive positive advantages, even when their absolute rewards are reduced. As shown in \cref{fig:advantage}, this effect is most evident for low-$\alpha$ instances, where successful long trajectories retain positive optimization signals despite high token usage.
Thus, BWS does not merely reduce variance; it preserves global difficulty signals in the advantage function, enabling credit assignment for deliberate but necessary long-horizon reasoning.

\paragraph{GRPO: Removing Cross-Instance Reward Structure.}
In contrast, GRPO-style normalization computes advantages relative only to rollouts of the same problem, removing any cross-instance reward comparisons. While this eliminates instance-level scale variation, it also discards the global difficulty structure encoded in IDAC.
As a consequence, the advantage landscape becomes fragmented, manifesting as horizontal bands in \cref{fig:advantage}. Because normalization is restricted to within-problem rollouts, credit assignment becomes sensitive to the composition of sampled trajectories. In particular, long but successful reasoning traces on difficult problems may receive weak or unstable signals when competing against short incorrect or prematurely terminated rollouts.
These results highlight a key limitation of instance-local normalization, and motivate BWS as a mechanism for preserving cross-instance difficulty structure in critic-free optimization.

\clearpage
\subsection{Analysis of BWS Training Dynamics}
\label{appx:abl_bws}

\begin{figure}[H]
    \centering
    \includegraphics[width=\textwidth]{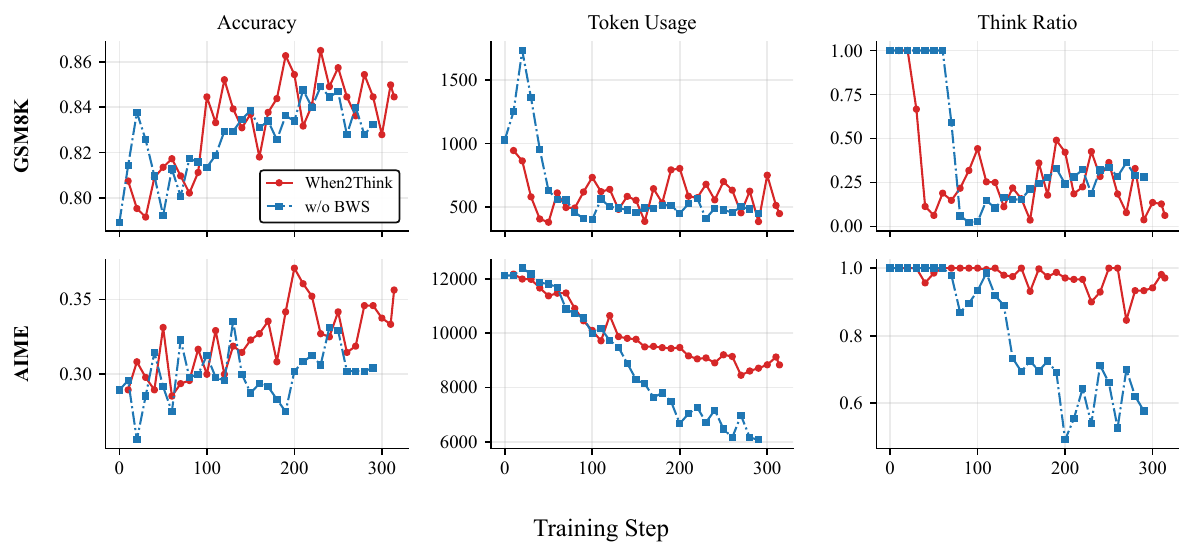}
    \caption{\textbf{Training Dynamics under Batch-Wise Standardization Ablations.} 
    \emph{When2Think} (\textcolor{BrickRed}{red}, with batch-wise standardized advantage) and \emph{Baseline} (\textcolor{NavyBlue}{blue}, without standardization) on DeepScaleR. 
    \textbf{Top:} GSM8K (easy). 
    \textbf{Bottom:} AIME24 (hard). 
    Each row reports validation accuracy, average token usage, and Think Ratio over training steps.}
    \label{fig:training_metrics}
\end{figure}

We analyze training dynamics to evaluate the effectiveness of our optimization strategy, focusing on the role of \textit{Batch-Wise Standardization} (BWS).
\cref{fig:training_metrics} reports both task performance and reasoning behavior throughout post-training.

\paragraph{Improved Convergence and Stability.}
BWS substantially stabilizes policy optimization.
Without BWS, raw reward signals exhibit high variance across instances, leading to noisy gradient estimates and unstable learning trajectories.
By normalizing rewards using batch-level statistics ($\mu_k, \sigma_k$), BWS reduces gradient variance and disentangles instance difficulty from policy quality in the optimization signal.
This effect is particularly pronounced on AIME24, where problem difficulty varies sharply; under BWS, the model consistently improves performance even on the hardest instances.

\paragraph{Controlled Reasoning Depth.}
BWS also plays a critical role in regulating reasoning length.
Without proper normalization, baseline policies tend to drift toward degenerate behaviors--either collapsing to minimal reasoning or producing excessively verbose outputs--due to unstable credit assignment.
In contrast, \emph{When2Think} maintains a stable \textit{Think Ratio} and \textit{Response Length}, gradually adapting reasoning depth to optimize the accuracy--efficiency trade-off.
Notably, this behavior emerges without relying on an explicit value function, indicating that BWS effectively stabilizes critic-free policy optimization.

\paragraph{Sample Efficiency and Difficulty Scaling.}
Our approach exhibits improved sample efficiency.
On AIME24, \emph{When2Think} reaches peak performance in approximately half the training steps required by the baseline.
Moreover, the performance gap widens as task difficulty increases, highlighting robustness under high-entropy reasoning regimes.
Finally, BWS decouples response length from accuracy, preventing the reasoning bloat observed in the baseline, where additional tokens fail to yield meaningful performance gains.

\clearpage
\subsection{Analysis of IS \& IDAC Training Dynamics}
\label{appx:abl_is_IDAC}

We further analyze the roles of Importance Sampling (IS) and Instance-level Difficulty-Aware Control (IDAC) during training. We compare three variants: the full \emph{When2Think} model, a variant without IS (retaining IDAC and BWS), and a variant without IDAC (retaining IS and BWS). 
This ablation isolates whether the observed accuracy--efficiency trade-offs are driven by balanced \textsc{Think}/\textsc{NoThink} exploration (via IS) or by difficulty-aware reward shaping and normalized optimization signals (via IDAC and BWS).

\begin{figure}[H]
    \centering
    \includegraphics[width=\textwidth]{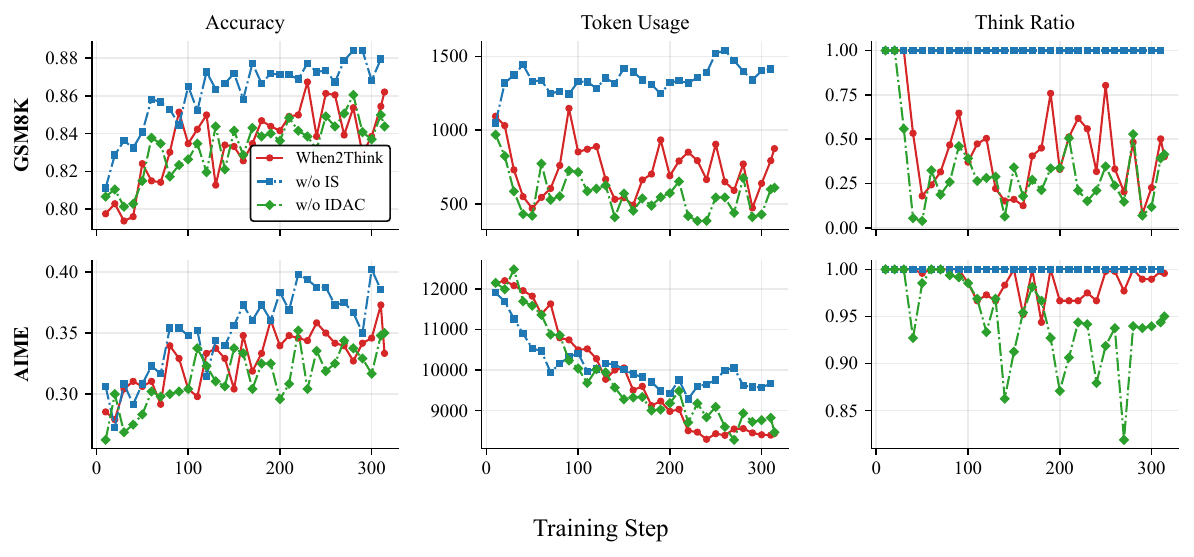}
    \caption{\textbf{Training Dynamics under IS, IDAC Ablations.}
    Training dynamics of three variants on DeepScaleR: the full \emph{When2Think} model
    (\textcolor{BrickRed}{IS + IDAC + BWS}), a variant without IS
    (\textcolor{NavyBlue}{IDAC + BWS}), and a variant without IDAC
    (\textcolor{ForestGreen}{IS + BWS}).
    \textbf{Top:} GSM8K (easy). 
    \textbf{Bottom:} AIME24 (hard). 
    Each row reports validation accuracy, token usage, and Think Ratio over training steps.}
    \label{fig:is_dald}
\end{figure}

\paragraph{IDAC and BWS provide the main depth-control signal.}
The \emph{IDAC+BWS} variant removes the balanced exploration distribution induced by IS, while retaining difficulty-aware reward shaping and batch-wise standardized advantages.
As shown in \cref{fig:is_dald}, this variant remains competitive in validation accuracy, especially on the harder AIME24 benchmark.
This suggests that the main depth-control signal is primarily driven by IDAC-based reward shaping together with BWS-based advantage standardization, rather than by Importance Sampling alone.

\paragraph{IS shifts the policy toward a more efficient hybrid operating point.}
Although IS is not the primary source of the depth-control signal, it changes the operating point of the learned policy.
On the easier GSM8K benchmark, the full \emph{When2Think} model achieves comparable validation accuracy while reducing both token usage and Think Ratio relative to the \emph{IDAC+BWS} variant.
This indicates that IS helps expose and exploit the \textsc{NoThink} mode during training, encouraging more efficient behavior when explicit reasoning is unnecessary.

\paragraph{IDAC is necessary for difficulty-aware depth control.}
The \emph{IS+BWS} variant retains balanced mode exploration but removes IDAC.
While this variant can still learn to switch between \textsc{Think} and \textsc{NoThink}, it lacks the difficulty-aware signal that determines how much reasoning should be allocated once \textsc{Think} is selected.
As a result, IS alone can encourage mode switching, but does not provide the same control over reasoning depth across easy and hard instances.

\clearpage
\subsection{Ablation on Hybrid Bonus Design}
\label{appx:bonus_design}

\begin{figure}[H]
    \centering
    \includegraphics[width=\textwidth]{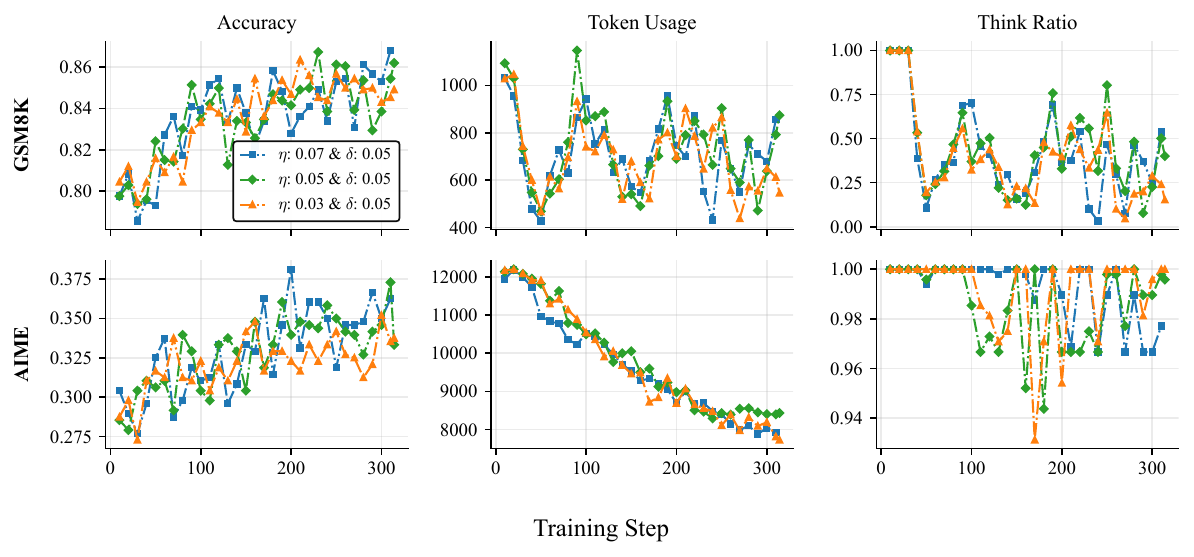}
    \caption{\textbf{Training Dynamics under $\eta$ and $\delta$ Variations.}
    Training dynamics of \emph{When2Think} under different $\eta$ on DeepScaleR, with $\delta$ fixed to 0.05.
    \textbf{Top:} GSM8K (easy).
    \textbf{Bottom:} AIME24 (hard).
    Each row reports validation accuracy, token usage, and Think Ratio over training steps.
    }
    \label{fig:eta_delta}
\end{figure}

\paragraph{Separate bonuses provide limited benefit.}
As shown in \cref{fig:eta_delta}, varying the \textsc{NoThink} bonus $\eta$ does not yield a clear improvement in the accuracy--efficiency trade-off.
In particular, configurations with larger $\eta$ tend to produce lower validation accuracy, suggesting that overly strong incentives for \textsc{NoThink} can bias the policy toward direct answering even when explicit reasoning remains useful.
This behavior is consistent with hard-instance under-allocation: encouraging \textsc{NoThink} too aggressively can suppress necessary long-horizon reasoning.

\begin{figure}[H]
    \centering
    \includegraphics[width=\textwidth]{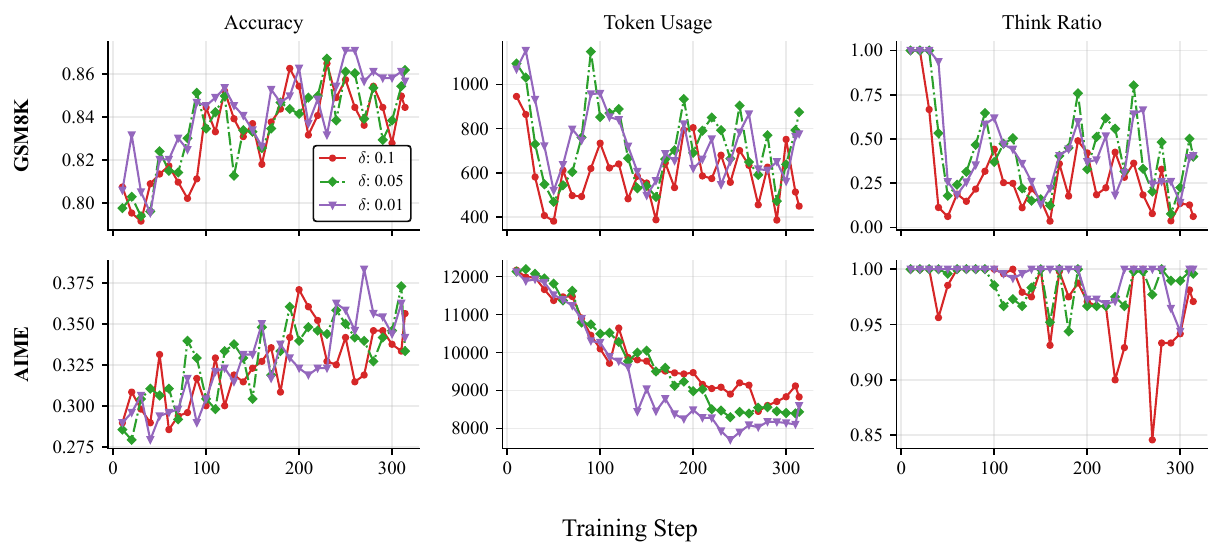}
    \caption{\textbf{Training Dynamics under $\delta$ Variations.}
    Training dynamics of \emph{When2Think} under different efficiency-bonus coefficients $\delta$ on DeepScaleR.
    \textbf{Top:} GSM8K (easy).
    \textbf{Bottom:} AIME24 (hard).
    Each row reports validation accuracy, token usage, and Think Ratio over training steps.}
    \label{fig:delta}
\end{figure}

\paragraph{Unified bonus yields more stable training.}
We therefore simplify the reward by using a single correctness-gated efficiency coefficient $\delta$ for both modes, with IDAC controlling how strongly the bonus is applied to \textsc{Think} trajectories.
The training dynamics in \cref{fig:delta} show that smaller unified bonus values lead to more stable behavior, avoiding both excessive \textsc{NoThink} collapse and overly verbose \textsc{Think} trajectories.
This suggests that the reward does not require separate mode-specific bonus coefficients.

\clearpage
\subsection{Effectiveness of the Dual Math Verifier}
\begin{figure}[H]
    \centering
    \begin{subfigure}[h]{0.32\textwidth}
        \centering
        \includegraphics[width=\linewidth]{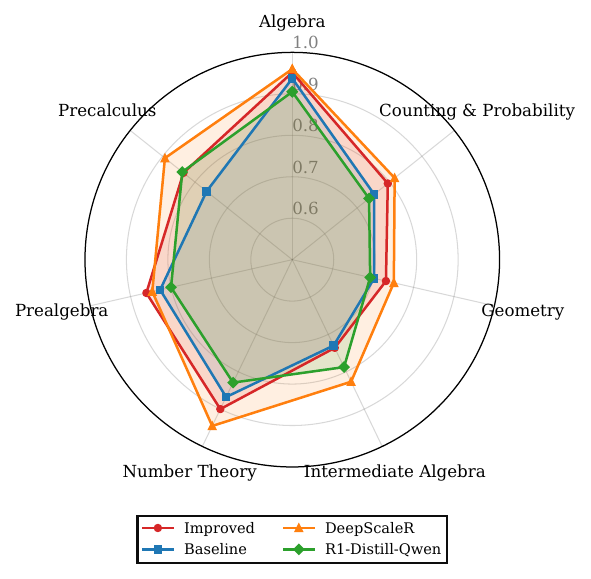}
        \caption{Performance Comparison}
    \end{subfigure}
    \hfill
    \begin{subfigure}[h]{0.32\textwidth}
        \centering
        \includegraphics[width=\linewidth]{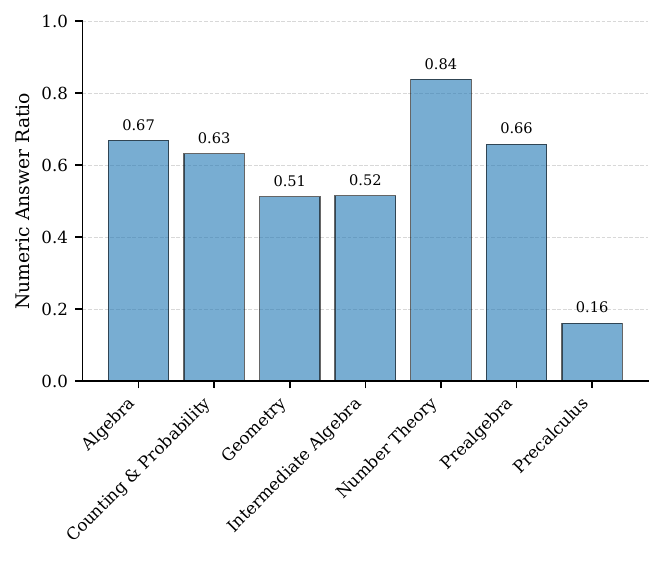}
        \caption{Ratio of Numeric Answers}
    \end{subfigure}
    \hfill
    \begin{subfigure}[h]{0.32\textwidth}
        \centering
        \includegraphics[width=\linewidth]{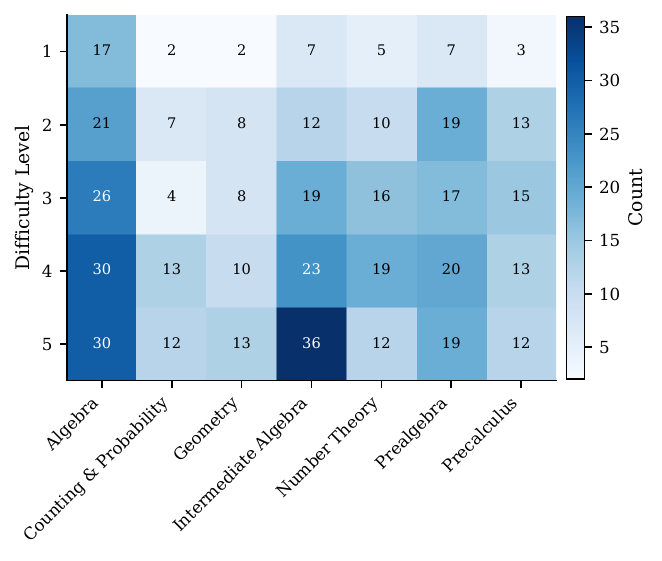}
        \caption{Distribution of Problems}
    \end{subfigure}
    \caption{\textbf{Subject-wise analysis of MATH-500.} 
    \textbf{(a)} Performance across subjects. The proposed method (Improved) recovers baseline drops in geometry and precalculus. 
    \textbf{(b)} Fraction of numeric-answer problems; low numeric ratios align with baseline failures. 
    \textbf{(c)} Heatmap of problem counts across difficulty levels (1--5), confirming sufficient sample coverage for each subject.}
    \label{fig:math500_analysis}
\end{figure}

We analyze the contribution of the Dual Math Verifier through subject-level performance on MATH-500.
\cref{fig:math500_analysis} compares the Base Model (R1-Distill), the Baseline with standard verification, and our proposed method (Improved).
\cref{fig:math500_analysis}(c) further confirms that each subject exhibits sufficient coverage in both problem count and difficulty distribution, enabling meaningful comparison.

\paragraph{Robustness to Non-Numeric Answers.}
Subjects such as \textit{Geometry} and \textit{Precalculus} contain a high proportion of symbolic, structured, or text-heavy answers.
In these categories, the Baseline exhibits notable performance degradation due to the inherent limitations of string-based verification, which fails to recognize semantically equivalent but syntactically distinct solutions (e.g., reordered vector coordinates or algebraically equivalent expressions).
By incorporating symbolic equivalence checking, the Dual Math Verifier effectively resolves this mismatch, substantially improving robustness to non-numeric answer formats and yielding consistent gains across these subjects.

\paragraph{Trade-offs in Intermediate Algebra.}
Performance gains in \textit{Intermediate Algebra} are comparatively smaller.
This behavior is not attributable to data scarcity, as models such as DeepScaleR--trained on the same dataset--achieve stronger results, but rather reflects a trade-off between verification strictness and recall.
In particular, complex multi-step derivations involving alternative but equivalent forms (e.g., factored versus expanded polynomials) may be falsely rejected under strict symbolic parsing.
Moreover, efficiency constraints imposed by IDAC can limit the allocation of reasoning steps required for such transformations.
These observations suggest that subject-aware verification tolerance or adaptive length modulation may further improve performance in algebra-intensive domains.

\clearpage
\section{Case Studies}

\subsection{Overthinking on Easy Problem}

\begin{figure}[H]
    \centering
    \begin{tcolorbox}[title={[GSM-Plus; R1-Distill-Qwen-1.5B] Correct Overthinking Response (\num{32206} tokens)}, colframe=YellowGreen, coltitle=white]
        \input{assets/overthink_correct_r1-1-5b}
    \end{tcolorbox}
    \caption{R1-Distill-1.5B GSM-Plus correct overthinking}
\end{figure}

\clearpage

\subsection{Underthinking on Hard Problem}

\begin{figure}[H]
    \centering
    \begin{tcolorbox}[title={[AIME24; R1-Distill-Qwen-1.5B] Incorrect Underthinking Response (\num{17075} tokens)}, colframe=YellowOrange, coltitle=white]
        \input{assets/underthink_incorrect-r1}
    \end{tcolorbox}
    \caption{R1-Distill-1.5B GSM-Plus incorrect underthinking}
\end{figure}

\clearpage
\subsection{Comparison with When2Think and Base Model}

\begin{figure}[H]
    \centering
    \begin{tcolorbox}[enhanced, title={[GSM-Plus] Problem}]
      \textbf{\textit{Nate's dog can dig six holes a day. Each hole takes 15 minutes. He digs for 14 days while Nate is on vacation. When Nate gets home, he starts filling in 9 holes a day, but the dog keeps digging 6 new holes every night. How many weeks does it take him to fill in all the holes?}}
    \end{tcolorbox}
\end{figure}

\begin{figure}[H]
    \centering
    \begin{tcolorbox}[enhanced, title={[GSM-Plus; When2Think-1.5B] Correct \textsc{NoThink} Response (\num{625} tokens)}, colframe=ForestGreen, coltitle=white]
        \input{assets/overthinking-when2think}
    \end{tcolorbox}
    \caption{When2Think correct \textsc{NoThink}}
\end{figure}

\clearpage

\begin{figure}[H]
    \centering
    \begin{tcolorbox}[enhanced, title={[GSM-Plus; R1-Distill-Qwen-1-5B] Incorrect Response (\num{8837} tokens)}, colframe=BrickRed, coltitle=white]
        \input{assets/overthinking_incorrect-r1}
    \end{tcolorbox}
    \caption{R1-Distill-1.5B incorrect overthinking}
\end{figure}

\clearpage
\section{Problem Characterization}
\label{appx:prob}

\subsection{Impact of Model Scale and Supervision on Reasoning Efficiency}
\label{sec:scale_supervision}

\begin{figure}[H]
    \centering
    \begin{subfigure}[h]{0.32\textwidth}
        \centering
        \includegraphics[width=\linewidth]{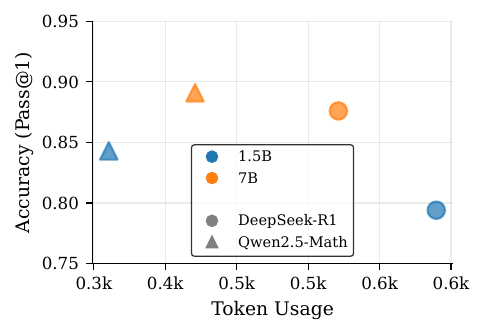}
        \caption{Qwen2.5-Math \& R1-Distill}
    \end{subfigure}
    \begin{subfigure}[h]{0.32\textwidth}
        \centering
        \includegraphics[width=\linewidth]{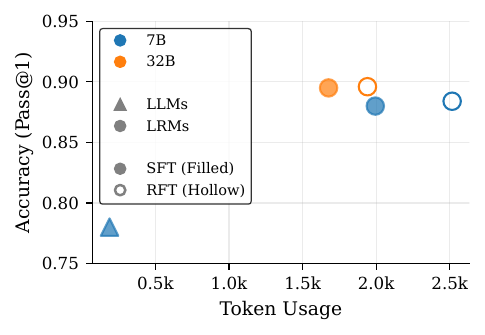}
        \caption{Olmo-3}
    \end{subfigure}
    \caption{\textbf{Efficiency landscape across model scales and tuning methods on GSM-Plus.}
    \textbf{(a) Model Scale and Reasoning Cost:} Reasoning-oriented models (LRMs, circles) consistently achieve higher accuracy than standard LLMs (triangles) but at the cost of significantly higher token usage. 
    Notably, smaller LRMs (1.5B, blue) consume more tokens than larger LRMs (7B, orange), suggesting that smaller models require longer reasoning chains to compensate for limited capacity.
    \textbf{(b) Impact of Tuning Strategies (SFT vs. RFT):} Across both model sizes, RFT models (hollow markers) are systematically shifted to the right compared to SFT models (filled markers), indicating that reinforcement fine-tuning induces higher computational overhead regardless of model scale.
    }
    \label{fig:problem_token}
\end{figure}

We examine reasoning efficiency across model scale and supervision using the Olmo-3~\citep{olmo2025olmo3} family (7B and 32B).
\cref{fig:problem_token} shows the accuracy--token usage trade-off across model scales, post-training strategies (SFT vs.\ RFT), and reasoning paradigms (standard LLMs vs.\ LRMs).
The results highlight systematic inefficiencies in current post-training approaches and illustrate how model capacity and supervision affect reasoning cost.

\paragraph{The Paradox of Scale: Smaller Models Overthink.}
\cref{fig:problem_token} reveals a clear \textit{paradox of scale} in reasoning efficiency.
Smaller 7B models are concentrated in the high token-usage regime relative to 32B models, despite achieving substantially lower accuracy.
This indicates that limited model capacity is associated with inefficient and prolonged reasoning trajectories, a phenomenon referred to as \textit{capability-induced overthinking}.
In contrast, 32B models achieve higher accuracy with fewer tokens, exhibiting superior \emph{reasoning economy}.
A similar trend is observed in the R1-distill family (\cref{tab:eval_math}), suggesting that stronger base capabilities enable more direct and efficient problem-solving paths.

\paragraph{The Cost of Reinforcement Learning.}
\cref{fig:problem_token} shows a consistent \emph{efficiency tax} under Reinforcement Fine-Tuning (RFT).
Within the same model scale, RFT models (hollow markers) are systematically shifted toward higher token usage compared to their Supervised Fine-Tuned (SFT) counterparts (filled markers).
These shifts are accompanied by only modest accuracy gains, resulting in a less favorable accuracy--token trade-off.
This behavior is consistent with outcome-based reinforcement learning objectives that do not explicitly penalize token usage, leading to longer chain-of-thought generations across instances.
Overall, RFT models occupy a Pareto-inferior region relative to SFT baselines.

\paragraph{The Mechanics of Inefficiency: Why RLVR Inflates Tokens.}
The token inflation observed in RFT models--particularly those trained with RLVR~\citep{lambert2025tulu}, such as OlmoRL~\citep{olmo2025olmo3}--follows directly from the optimization objective.
Unlike SFT, which imitates concise human-curated reasoning traces, RLVR optimizes only for outcome correctness and imposes no explicit constraint on token usage.
The absence of a KL regularization term further permits substantial deviation from the SFT initialization.
Consequently, longer reasoning trajectories become a favorable strategy for increasing success probability.
Without explicit difficulty-aware control, RLVR therefore tends to produce inefficient and overly verbose reasoning behavior.

\clearpage

\end{document}